\documentclass{article}

\usepackage{microtype}
\usepackage{graphicx}
\usepackage{subfigure}
\usepackage{booktabs} 

\usepackage{hyperref}

\usepackage[accepted]{icml2025}

\usepackage{amsmath}
\usepackage{amssymb}
\usepackage{mathtools}
\usepackage{amsthm}

\usepackage[capitalize,noabbrev]{cleveref}

\theoremstyle{plain}

\theoremstyle{definition}

\theoremstyle{remark}

\usepackage[textsize=tiny]{todonotes}

\usepackage[utf8]{inputenc} 
\usepackage[T1]{fontenc}    
\usepackage{hyperref}       
\usepackage{url}            
\usepackage{booktabs}       
\usepackage{amsfonts}       
\usepackage{nicefrac}       
\usepackage{microtype}      
\usepackage{xcolor}         
\usepackage{longtable}

\usepackage{amsmath,amsfonts,bm}

\def\eqref#1{equation~\ref{#1}}

\def\1{\bm{1}}

\DeclareMathAlphabet{\mathsfit}{\encodingdefault}{\sfdefault}{m}{sl}
\SetMathAlphabet{\mathsfit}{bold}{\encodingdefault}{\sfdefault}{bx}{n}

\def\gA{{\mathcal{A}}}

\def\gO{{\mathcal{O}}}

\def\gQ{{\mathcal{Q}}}

\DeclareMathOperator*{\argmax}{arg\,max}

\usepackage{xspace}
\usepackage{xcolor}
\usepackage{pifont}
\usepackage{tcolorbox}

\newcommand{\cmark}{\ding{51}}  
\newcommand{\xmark}{\ding{55}}  

\newcommand{\ie}{\textit{i}.\textit{e}.}
\newcommand{\eg}{\textit{e}.\textit{g}.}

\newcommand{\etc}{\textit{etc}.}

\definecolor{lightgray}{rgb}{0.7, 0.7, 0.7}
\newcommand{\myalgorithmiccomment}[1]{\hfill $\triangleright$ \textcolor{gray}{#1}}

\newcommand{\name}{\texttt{LAC}}
\newcommand{\namewolangcritic}{\texttt{LAC} \textit{w/o} reflection}
\newcommand{\namewovaluecritic}{\texttt{LAC} \textit{w/o} critic}
\newcommand{\nameworollout}{\texttt{LAC} \textit{w/o} rollout}
\newcommand{\nameonlyvaluecritic}{\textit{critic-only}}
\newcommand{\react}{ReAct}
\newcommand{\icpi}{ICPI}
\newcommand{\rap}{RAP}
\newcommand{\rafa}{RAFA}
\newcommand{\lats}{LATS}
\newcommand{\alfred}{ALFRED}

\def\piL{\pi_{\text{LLM}}}
\def\piN{\pi_{\text{new}}}

\def\gVL{\gQ_{\text{LLM}}}
\def\fL{f_{\text{LLM}}}

\newcommand{\codellamaSeven}{CodeLlama-7B}
\newcommand{\mistralSeven}{Mistral-7B}
\newcommand{\gemmaSeven}{Gemma-7B}
\newcommand{\llamaEight}{Llama-3-8B}
\newcommand{\llamaThreeOneEight}{Llama-3.1-8B}
\newcommand{\llamaThreeOneEightIt}{Llama-3.1-8B-It}
\newcommand{\gptFour}{GPT-4}

\newcommand{\alfworld}{ALFWorld}
\newcommand{\babyai}{BabyAI-Text}
\newcommand{\webshop}{WebShop}
\newcommand{\balrog}{BALROG}
\newcommand{\crafter}{Crafter}

\definecolor{easycolor}{RGB}{45,106,65}
\definecolor{mediumcolor}{RGB}{223,140,67}
\definecolor{hardcolor}{RGB}{208,44,46}

\definecolor{lightgrey}{RGB}{211,211,211}
\definecolor{grey}{RGB}{130, 130, 130}
\definecolor{critic}{RGB}{104,52,154}
\definecolor{thought}{RGB}{79,173,91}

\icmltitlerunning{LLM-based Actor-Critic}

\begin{document}

\twocolumn[
\icmltitle{Enhancing Decision-Making of Large Language Models via Actor-Critic}



\icmlsetsymbol{equal}{*}


\begin{icmlauthorlist}
\icmlauthor{Heng Dong}{equal,thu}
\icmlauthor{Kefei Duan}{equal,wustl}
\icmlauthor{Chongjie Zhang}{wustl}
\end{icmlauthorlist}

\icmlaffiliation{thu}{IIIS, Tsinghua University}
\icmlaffiliation{wustl}{Washington University in St. Louis}

\icmlcorrespondingauthor{Heng Dong}{drdhxi@gmail.com}

\icmlkeywords{Machine Learning, ICML}

\vskip 0.3in
]



\printAffiliationsAndNotice{\icmlEqualContribution} 

\begin{abstract}
Large Language Models (LLMs) have achieved remarkable advancements in natural language processing tasks, yet they encounter challenges in complex decision-making scenarios that require long-term reasoning and alignment with high-level objectives. Existing methods either rely on short-term auto-regressive action generation or face limitations in accurately simulating rollouts and assessing outcomes, leading to sub-optimal decisions. This paper introduces a novel LLM-based Actor-Critic framework, termed \name{}, that effectively improves LLM policies with long-term action evaluations in a principled and scalable way. Our approach addresses two key challenges: (1) extracting robust action evaluations by computing Q-values via token logits associated with positive/negative outcomes, enhanced by future trajectory rollouts and reasoning; and (2) enabling efficient policy improvement through a gradient-free mechanism. Experiments across diverse environments -- including high-level decision-making (ALFWorld), low-level action spaces (BabyAI-Text), and large action spaces (WebShop) -- demonstrate the framework’s generality and superiority over state-of-the-art methods. Notably, our approach achieves competitive performance using 7B/8B parameter LLMs, even outperforming baseline methods employing GPT-4 in complex tasks. These results underscore the potential of integrating structured policy optimization with LLMs’ intrinsic knowledge to advance decision-making capabilities in multi-step environments.
\end{abstract}

\section{Introduction}

Large Language Models (LLMs) \citep{touvron2023llama2, jiang2023mistral, team2024gemma} have demonstrated impressive capabilities across various natural language processing tasks, including text generation \citep{10.1145/3649449}, question answering \citep{li2024unigen}, and summarization \citep{jin2024comprehensive}. The successful application of LLMs in these areas, coupled with their extensive internal knowledge, has generated significant interest in leveraging LLMs to tackle complex decision-making problems, particularly in data-scarce environments.

Early works \citep{ahn2022saycan, huang2022language} have employed LLMs as policies, generating actions in an auto-regressive way that directly utilizes the models' prior knowledge for decision-making. While these methods are simple and effective for short-term action generation, they often lack the capacity for long-term planning. Although reasoning techniques such as Chain-of-Thought have been introduced to enhance the reasoning capabilities of LLMs \citep{yao2022react, shinn2024reflexion} and improve action selection, the ability to engage in comprehensive long-term planning remains underdeveloped. Consequently, decisions may appear locally optimal but fail to meet the overall objectives in more complex, multi-step environments.

Other approaches \citep{hao2023rap, liu2023rafa, fu2024preact, brooks2024large} incorporate planning, either through interaction with real environments or by leveraging the LLMs' internal imaginative capabilities, followed by action evaluation based on the planning outcomes. These evaluations are then used to select the actions that yield the best results. However, such methods heavily depend on the model's accuracy in performing rollouts and evaluating outcomes, leading to sub-optimal action selection when the model's rollouts diverge from reality or when action evaluations are inaccurate.

\begin{figure*}[t]
\centering
\includegraphics[width=\linewidth]{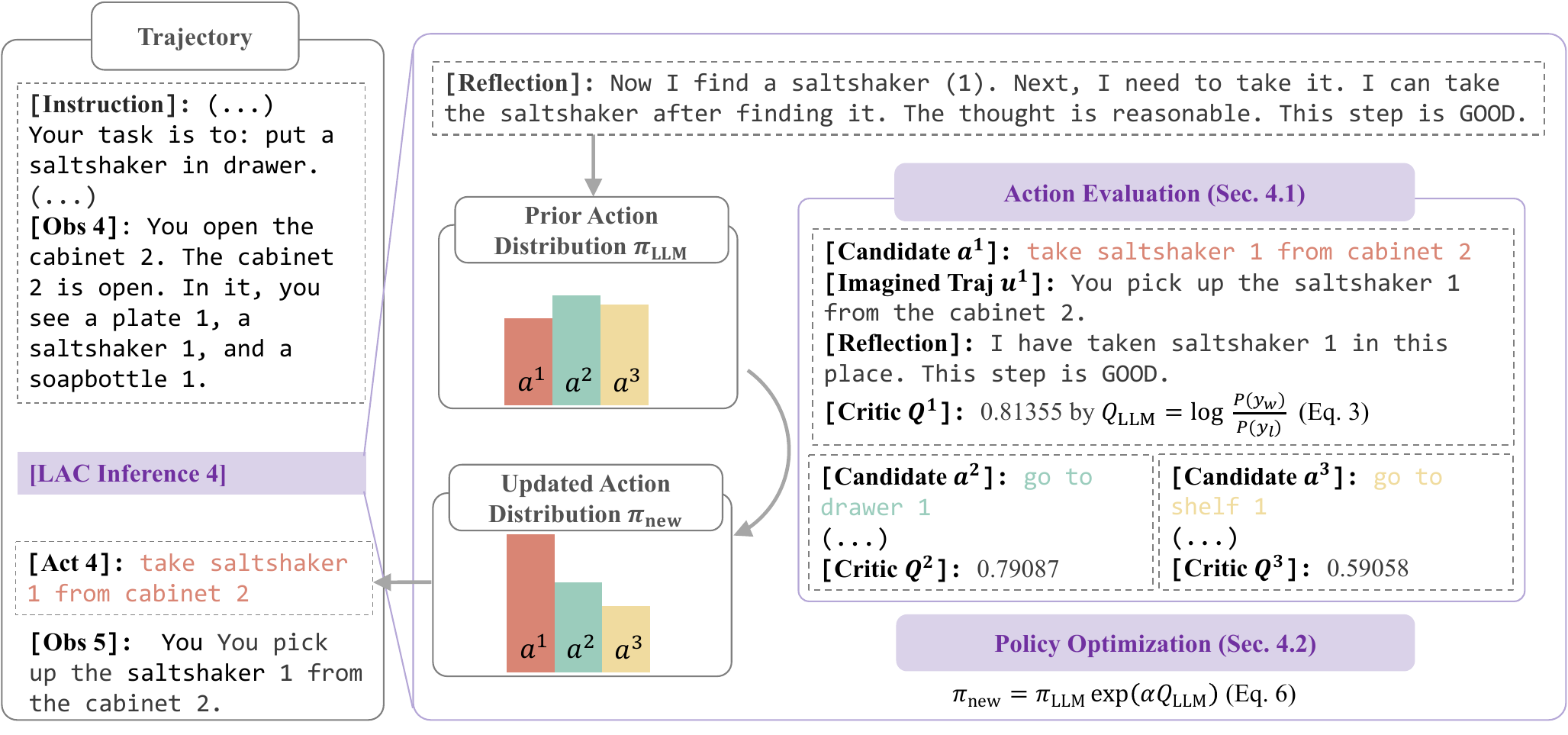}
\vspace{-1em}
\caption{Framework of our \name{}. At each time step, \name{} optimizes the policy via two steps: (1) the critic $\gVL$ evaluates each candidate action sampled from the policy $\piL$; (2) the policy $\piL$ is optimized according to the action evaluations using a gradient-free method.}\label{fig:framework}
\end{figure*}

The fundamental limitation of these two lines of research lies in the decoupling of LLM's prior policy and action evaluations. They either (1) rigidly follow LLM priors without sufficient planning or (2) over-rely on potentially flawed simulated rollouts, both fail to reconcile the LLM's inherent knowledge with error-corrected planning insights. 

To address the above problems, we propose a novel LLM-based Actor-Critic (\name{}) approach that leverages action evaluations to optimize the LLM's prior policy under constraints, facilitating the effective integration of decision-making information inherent in the LLM's prior knowledge with insights derived from robust action evaluations that incorporate long-term planning. While there have been attempts to simultaneously utilize both types of information (the LLM's prior knowledge and action evaluations) for decision-making, these efforts have typically resulted in simplistic combinations \citep{zhang2023controlling}.

Implementing this idea involves two major challenges. First, how to extract action evaluation information from the LLM. A simple approach is to directly prompt the LLM to output evaluation values that indicate the quality of action; however, this direct prompting method can be unstable and easily influenced by the instructions and examples provided to LLMs. To address this issue, we use the logits of specific tokens to calculate Q values (action values). Specifically, we employ tokens that convey positive or negative meanings as the LLM's internal beliefs about the action's quality or its likelihood of successfully completing the task. We derived a simple yet effective formula to use these logits for Q value computation. Additionally, to allow long-term planning ability and enhance the accuracy of this calculation, we perform rollouts for each candidate action to predict future trajectories, conduct a brief reasoning process based on these trajectories, and then collect positive/negative token logits to compute Q values. 

The second challenge is how to utilize these action evaluation insights to optimize the policy. A straightforward method might involve direct gradient-based actor-critic updates; however, this approach is inefficient for LLMs due to high computational demands, leading to increased decision latency. To solve this challenge, we formulate the policy improvement problem as a KL-divergence optimization problem and derive a closed-form solution, allowing us to optimize the policy in a gradient-free manner and improve the accuracy of decision-making.

Empirically, we demonstrate the effectiveness and generality of \name{} across diverse environments, including high-level decision-making tasks (\alfworld{}, \citep{shridhar2021alfworld}), low-level action space (\babyai{}, \citep{carta2023grounding}) and potentially infinite action space (\webshop{}, \citep{yao2022webshop}). The results show that our approach consistently outperforms state-of-the-art methods, \eg{}, \rap{} \citep{hao2023rap}, \lats{} \citep{zhou2024lats}. Notably, in several challenging tasks, \name{} using 7B/8B LLMs significantly surpasses \react{} \cite{yao2022react} with \gptFour{} \citep{achiam2023gpt4}. 

Our contributions are twofold: (1) We propose a novel Q-function estimation approach to extract action evaluation information from LLMs that leverages LLMs' internal belief about success or failure of the current task; (2) We formulate the policy improvement problem as a KL-divergence constrained optimization and derive a closed-form solution, allowing us to optimize the policy in a gradient-free manner using the action evaluation.
\section{Related Work}

\paragraph{Large Language Models for Sequential Decision-Making}

Sequential decision-making is a key ability of intelligent agents, involving generating actions to achieve goals \citep{barto1989learning, littman1996algorithms, mccarthy1963situations, bylander1994computational}. Recently, LLM-based agents have gained popularity in decision-making across various fields, requiring only instructions or few-shot examples to adapt to new tasks \citep{huang2022monologue, singh2023progprompt, ding2023task} due to pre-training on large datasets. Previous work primarily categorizes LLMs into two roles: policies, which generate actions from trajectories, and critics, which evaluate actions based on trajectories and actions. Research falls into two categories: the first uses LLM-generated actions directly from previous trajectories \citep{ahn2022saycan, huang2022monologue, yao2022react, huang2022language, shinn2024reflexion}. However, the auto-regressive nature of LLMs limits their long-term planning capabilities, making them struggle with complex tasks \citep{huang2022survey, mialon2023survey}. The second category employs another LLM to evaluate actions by simulating outcomes, choosing the best one \citep{hao2023rap, liu2023rafa, fu2024preact, koh2024tree}. While these methods enable long-term planning, they heavily depend on evaluation accuracy, which can lead to sub-optimal solutions.

To address these issues, we propose optimizing LLM's prior policy using action evaluations under constraints, enhancing long-term planning while mitigating evaluation inaccuracies. Previous attempts to combine LLM's prior policy and action evaluation \citep{zhang2023controlling} have been simplistic and lacked theoretical guarantees, resulting in unsatisfactory performance.

\paragraph{Large Language Models with Reinforcement Learning}

Classical sequential decision-making methods, such as Reinforcement Learning (RL), have been widely adopted in embodied environments \citep{schulman2017PPO, fujimoto2018TD3, huang2020one, dong2022solar}. However, these RL-based methods are typically sample-inefficient and require lots of samples for training. On the other hand, LLMs that contain rich prior knowledge about the world may alleviate this burden. To combine RL and LLM, one straightforward way is to use LLMs as base models and add policy/value heads on top of LLMs \citep{carta2023glam, tan2024twosome}. Then classical RL methods like PPO \citep{schulman2017PPO} can be used for training \citep{szot2023large, zhouarcher}. However, these methods still require lots of training samples of the same tasks, which reduces the benefits of using LLM to some extent and contradicts our settings. There are also other paradigms for combining. RLEM \citep{zhang2024rlem} adopts Q-learning \citep{watkins1992qlearning} and an experience memory to update policies, but it may get stuck in the tasks with extremely sparse rewards like \alfworld{} and \babyai{}. Retroformer \citep{yao2023retroformer} trains a smaller LLM with PPO to generate suitable prompts for a larger LLM for a specific task, while our method only needs a small model. \icpi{} \citep{brooks2024large} uses LLMs to implement policy iteration by predicting future trajectories and accumulating future rewards, which may also struggle with sparse reward settings. We have compared it empirically in \cref{sec:exp}.
\section{Preliminary \& Previous Work}\label{sec:prelim}

In this section, we describe the task setting and previous LLM decision-making methods for better understanding. 

\textbf{Task setup.}
Consider a general setup of an agent interacting with an environment for achieving a given goal $g$, \eg{}, $g=$``put a clean egg in microwave'' (from \alfworld{}) or $g=$``pick up the green ball'' (from \babyai{}). At time step $t$, the agent receives a natural language described observation $o_t\in\gO$ from the environment. The agent then takes an action $a_t\in \gA$ sampled from policy $\pi(a|g,h_t)$, where $h_t:=(o_1,a_1,o_2,a_2\cdots,o_t)$ is the history to the agent. During execution, there is no immediate reward and only at the end of each episode, the environment will give a signal to evaluate the completion of the task. For each testing task, the agent can only try once and cannot conduct improvements through repeated trials. 

\textbf{Methods that directly use LLM's prior as a policy.} To solve the above tasks with LLMs, one simple method is to directly use LLM's prior as a policy: $a_t\gets \argmax_a \piL(a|g,h_t)$, which can be implemented by simply injecting instructions or few-show examples to the prompt as suggested in \citet{yao2022react}. Despite its simplicity, the policy $\piL$ generates actions solely relying on its auto-regression ability and it does not conduct long-term planning explicitly, which is typically necessary for sequential decision-making tasks. Additionally, this issue will be exacerbated when using lightweight models like \codellamaSeven{} \citep{roziere2023codellama} and \mistralSeven{} \citep{jiang2023mistral}. This problem is verified in \cref{sec:exp}.



\textbf{Methods that incorporate planning and action evaluations.} To handle the issue of lack of long-term planning, another line of research incorporates planning and action evaluations into decision-making \citep{hao2023rap, liu2023rafa, fu2024preact}. The basic idea is to first sample several candidate actions from policy $\{a_t^1,a_t^2,\cdots,a_t^n\}\sim \piL(\cdot|g,h_t)$, then evaluate each candidate action by other LLMs and finally select the action with the highest evaluation value. The evaluation procedure is the key to these methods, and many approaches can be adopted. For example, directly ask an LLM to evaluate the action candidate \citep{fu2024preact}, or predict the future trajectory $u_t$ of each action candidate by regarding an LLM as a forward world model $\fL$ and use the future outcome as evaluations, or use tree-search methods like Monte Carlo Tree Search (MCTS) \citep{kocsis2006bandit,coulom2006efficient} to expand each action candidate \citep{hao2023rap}. Despite this progress, these approaches heavily depend on the model's accuracy in performing rollouts and evaluating outcomes. When rollouts diverge from reality and evaluations are inaccurate, which could be common when using lightweight LLMs, the action selection could be sub-optimal.


\section{Method}\label{sec:method}



In this section, we introduce our novel LLM-based Actor-Critic (\name{}) algorithm, which effectively integrates the generative capabilities of LLMs (as the \textit{actor}) with their evaluative reasoning capabilities (as the \textit{critic}) in a principled and scalable manner. The actor, represented by the LLM-based prior policy ($\piL$), generates potential actions, while the critic ($\gVL$) evaluates these actions by incorporating long-term reasoning. This synergy enables effective policy improvement through an innovative gradient-free optimization method.

We first explain how to leverage LLMs to extract meaningful action evaluation information (\cref{sec:action_eval}). Building on this, we propose a gradient-free policy optimization method that efficiently refines the policy based on the critic's feedback (\cref{sec:policy_opt}). The overall workflow of the algorithm is summarized in \cref{alg:main}.





\begin{algorithm}[t]
\caption{\name{}: LLM-based Actor-Critic algorithm.}
\label{alg:main}
\begin{algorithmic}[1]
\STATE {\bfseries Input: }{current task goal $g$, history $h_t$, actor $\piL$, forward model $\fL$, value-based critic $\gVL$, hyperparameter $\alpha$, candidate action size $n$.}
\STATE {\bfseries Output: }{selected action $a_t^*$}
\STATE $\{a_t^i\}_{i=1}^n\sim \piL(\cdot|g,h_t)$; \myalgorithmiccomment{candidate actions}\\
\FOR{$i=1$ {\bfseries to} $n$}
    \STATE $u^i_t\leftarrow \fL(g,h_t,a^i_t)$; \myalgorithmiccomment{predict future trajectory}\\
    \STATE \textcolor{violet}{$\gVL(g,h_t,a_t^i,u_t^i) \gets \log \frac{P(y_w|g,h_t,a_t^i,u_t^i)}{P(y_l|g,h_t,a_t^i,u_t^i)}$}; \myalgorithmiccomment{{action evaluation (Sec. \ref{sec:action_eval})}}\\ 
\ENDFOR
\STATE \textcolor{violet}{$\piN(a_t^i|g,h_t)\gets \piL(a_t^i|g,h_t)\exp(\alpha \gVL(g,h_t,$\\$a_t^i,u_t^i))$}; \myalgorithmiccomment{{policy optimization (Sec. \ref{sec:policy_opt})}}\\ 
\STATE $a_t^* \gets \argmax_{a^i_t} \piN (a_t^i|g,h_t)$ \\
\end{algorithmic}
\end{algorithm}


\subsection{Critic Development for Long-term Action Evaluation}\label{sec:action_eval}


We propose a novel method for constructing the critic $\gVL$, designed to provide numerical evaluations of candidate actions sampled from the policy $\piL(\cdot|g,h_t)$. This approach targets goal-based decision-making problems characterized by sparse rewards and binary outcomes, where the agent receives a reward only upon achieving the goal. By linking $\gVL$ to the agent's success probability of task completion, we enable more effective guidance toward maximizing expected returns. Furthermore, we demonstrate how this evaluation can be derived directly from token logits associated with positive and negative outcomes.




\subsubsection{Connect {Critic} to Success Probability}

\begin{figure*}[t]
\centering
\includegraphics[width=\linewidth]{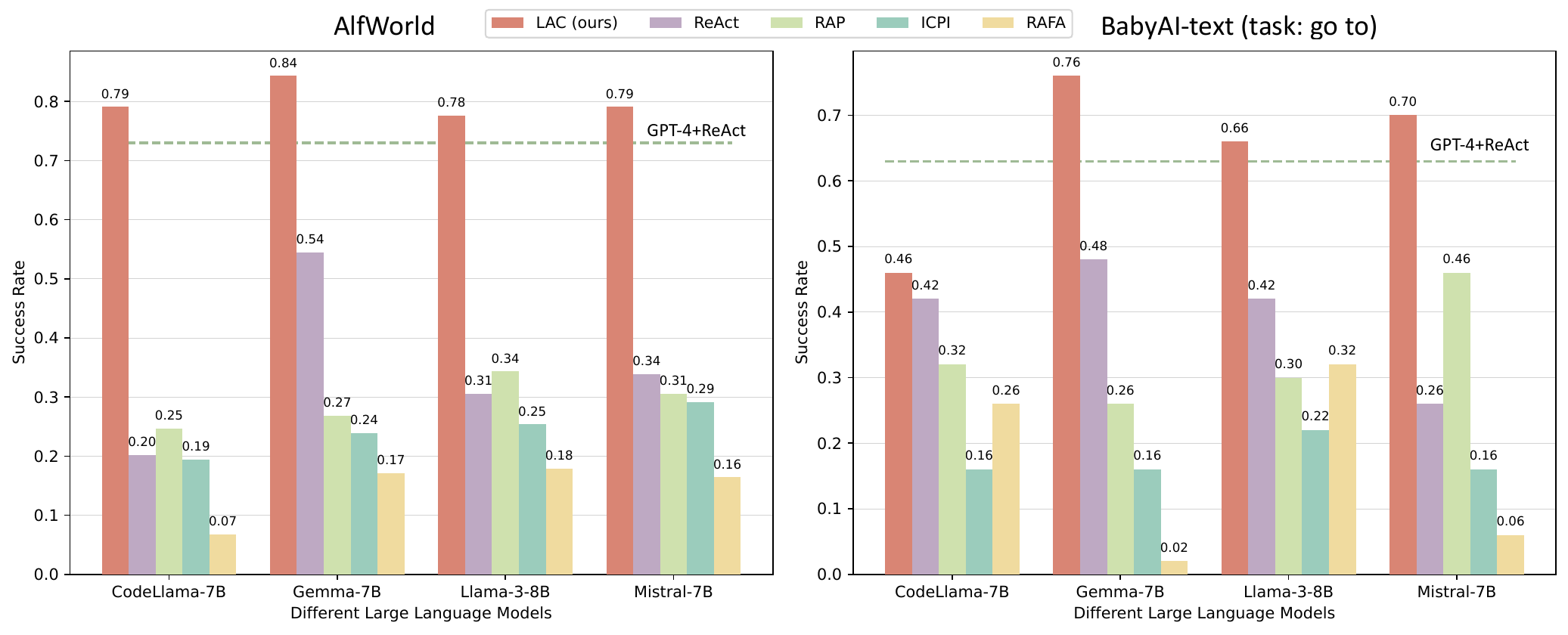}
\caption{Performance of our \name{} compared with various baselines in benchmarks \alfworld{} and \babyai{}.}\label{fig:performance}
\end{figure*}

Let $\gVL(g,h_t,a_t^i)$ be the value-based evaluation of each candidate action $a_t^i$ given the task goal $g$ and history $h_t$. We consider scenarios with sparse rewards, which are only provided at the end of each episode. 
Considering the binary outcomes, we hope $\gVL(g,h_t,a_t^i)$ could reflect the possibility of achieving success effectively. We employ a logistics function \citep{jordan1995logistic} to relate $\gVL(g,h_t,a_t^i)$ to the success probability.

Let $P(y_w|g,h_t,a_t^i) \in [0,1]$ denote the probability of successfully completing the task goal $g$ after executing action $a_t^i$, where $y_w$ represents a success signal at the end of the episode. Similarly, let $P(y_l|g,h_t,a_t^i)$ represent the failure probability. We use the following formulation to connect $P(y_w|g,h_t,a_t^i)$ with $\gVL(g,h_t,a_t^i)$: 
\begin{equation}
    P(y_w|g,h_t,a_t^i) = \frac{1}{1+\exp{(-\gVL(g,h_t,a_t^i))}}.\label{equ:logistic}
\end{equation}
With this formulation, $\gVL(g,h_t,a_t^i)$ is positively correlated with the success probability $P(y_w|g,h_t,a_t^i)$. Higher $\gVL(g,h_t,a_t^i)$ values map to a greater likelihood of success, allowing the critic to guide the policy toward actions that maximize long-term returns. In this way, maximizing the Q-function corresponds to maximizing the success probability for the current task.

While other formulations could be used, we found that \cref{equ:logistic} is both simple and effective for a wide range of tasks. For a comparison of alternative formulations, refer to \cref{appx:ablation-Q} and \cref{appx:direct_eval}.



\begin{figure*}[t]
\centering
\includegraphics[width=\linewidth]{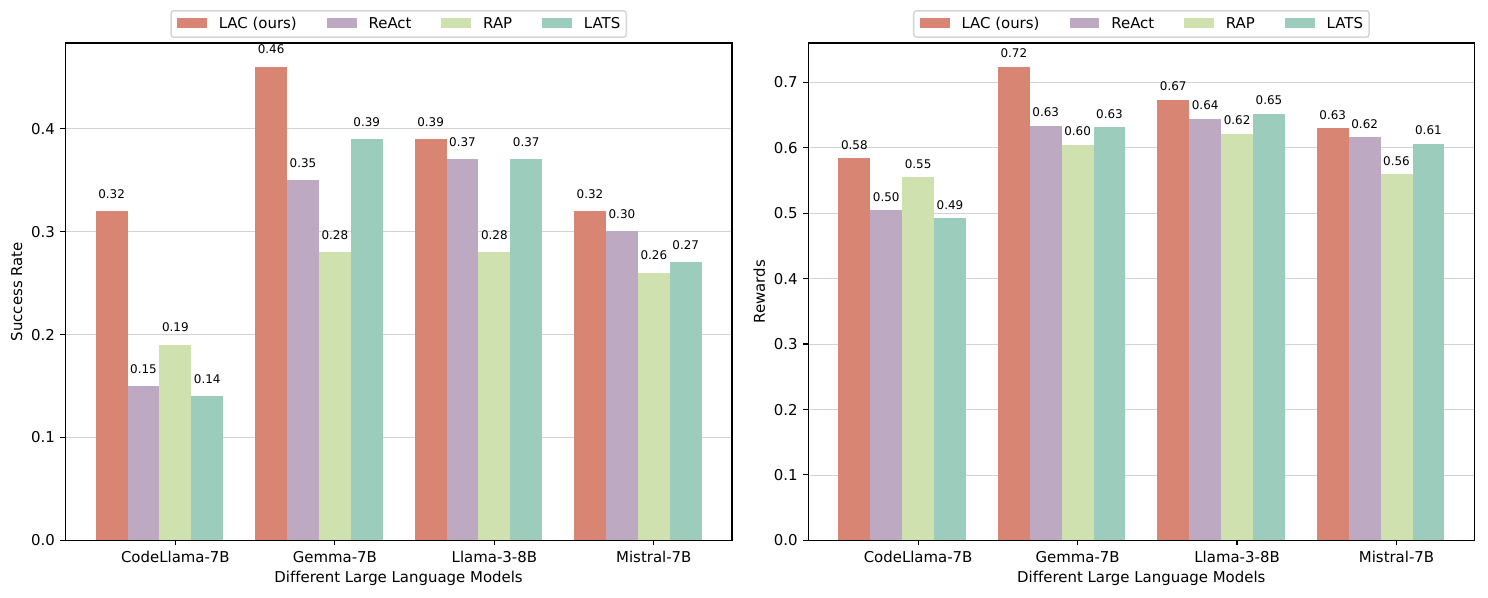}
\caption{Performance of \name{} in benchmark \webshop{}}\label{fig:exp-webshop}
\end{figure*}

\subsubsection{Estimate {Critic} with LLMs}
To estimate $\gVL(g,h_t,a_t^i)$ using LLMs, we perform an equivalent transformation on \cref{equ:logistic}:
\begin{equation}
    \gVL(g,h_t,a_t^i) = \log \frac{P(y_w|g,h_t,a_t^i)}{P(y_l|g,h_t,a_t^i)},\label{equ:def_q}
\end{equation}
which uses the equation $P(y_w)+P(y_l)=1$. With \cref{equ:def_q}, we can use the LLM to obtain $\gVL(g,h_t,a_t^i)$ via first estimating $P(y_{\{w,l\}}|g,h_t,a_t^i)$. The basic idea is to prompt the LLM to predict the outcomes given the current trajectory $(g,h_t)$ and action $a_t^i$. Specifically, we use special paired tokens with positive/negative meanings, \ie{}, ``GOOD''/``BAD'' or ``SUCCESS''/``FAILURE'', to indicate success/failure outcomes. The corresponding generated probabilities of LLMs for those special tokens reflect LLMs' internal belief in success/failure after taking action $a_t^i$. We let the generated probabilities of ``GOOD'' and ``BAD'' represent $P(y_{w}|g,h_t,a_t^i)$ and $P(y_{l}|g,h_t,a_t^i)$ respectively. 
Finally, using \cref{equ:def_q}, we can calculate $\gVL(g,h_t,a_t^i)$ for action $a_t^i$. Though our action evaluation is designed for binary outcomes, we empirically show that this formulation is also effective in continuous-reward settings in \cref{sec:exp}.

To improve the accuracy of $\gVL(g,h_t,a_t^i)$, we introduce future trajectory rollouts using a forward world model $\fL$, which can be implemented by prompting LLMs, \eg{}, adding few-shot examples, or by fine-tuning on these examples. For each candidate action $a_t^i$, we rollout several future steps to predict the resulting trajectory $u_t^i \sim \fL(g,h_t,a_t^i)$. By considering the future trajectory $u_t^i$, we obtain more informed estimates of the success and failure probabilities $P(y_{\{w,l\}}|g,h_t,a_t^i,u_t^i)$. This approach accounts for the delayed consequences of actions and ensures that $\gVL(g,h_t,a_t^i,u_t^i)$ reflects the long-term value of each action:
\begin{align}
    \gVL(g,h_t,a_t^i,u_t^i) = & \log \frac{P(y_w|g,h_t,a_t^i,u_t^i)}{P(y_l|g,h_t,a_t^i,u_t^i)}. \label{equ:def_q_future}
\end{align}
Trajectory rollouts are especially important in tasks where the outcomes of actions may unfold over several steps. By simulating the future impact of actions, the critic provides a more accurate assessment, guiding the policy towards actions that maximize the probability of long-term success.

Empirically, we also found that contextual reflections on previous actions can be helpful to improve performance. Specifically, given the task goal $g$ and history $h_t$, which may contain the predicted future trajectory, we prompt LLMs to generate a short reflection such as ``I have found object-X. This step is GOOD'' or ``I should take object-X instead of object-Y first. This step is BAD.'' These reflections provide judgments about whether and why the previous actions were appropriate. For more examples of these reflections, please refer to \cref{tab:alfworld_ours_example} and \cref{tab:babyai_ours_example} of \cref{appx:method}. 

These reflections serve as simple reasons, akin to a Chain-of-Thought (CoT) \citep{wei2022cot, kojima2022zerocot,prystawski2024think}, allowing the policy to sample better candidate actions by avoiding past mistakes. These reflections also enable the critic to evaluate 
candidate actions with greater accuracy, ultimately enhancing decision-making performance. For the difference comparison between our reflection and CoT, please refer to \cref{appx:reflection_cot}.

\subsection{Gradient-free Policy Optimization}\label{sec:policy_opt}

In this subsection, we derive a gradient-free policy optimization method using the above estimated value-based evaluation. To effectively improve the LLM's prior policy, we propose to use the following KL-constrained policy optimization problem to maximize the expected value function:
\begin{align}
    \max_\pi & \mathbb{E}_{a_t^i\sim \pi(\cdot|g,h_t), u_t^i \sim \fL(g,h_t,a_t^i)}\left[\gVL(g,h_t,a_t^i,u_t^i)\right] \notag\\
    & -\frac{1}{\alpha} \mathbb{D}_{KL}\left[\pi(a_t^i|g,h_t)\| \piL(a_t^i|g,h_t)\right],\label{equ:optimization}
\end{align}
where $\alpha$ is a hyperparameter controlling the deviation from the original policy $\piL$. The KL-divergence term prevents the new policy $\piN$ from deviating too far from the original policy, balancing the policy's prior knowledge and the critic’s guidance.

Following prior work \citep{rafailov2024dpo,go2023aligning,peng2019advantage,jain2013learning,peters2007reinforcement}, we can show that the optimal solution to the KL-constrained maximization objective in \cref{equ:optimization} takes the following form:
\begin{align}
    \piN(a_t^i|g,h_t)=&\frac{1}{Z(g,h_t)}\piL(a_t^i|g,h_t) \notag\\ 
    &\cdot\exp{(\alpha \gVL(g,h_t,a_t^i,u_t^i))},\label{equ:solution}
\end{align}
where $Z(g,h_t)=\sum_{a_t^i} \piL(a_t^i|g,h_t)\exp(\alpha \gVL(g,h_t,$ $a_t^i,u_t^i))$ is the partition function. Please refer to \cref{appx:derive_solution} for a complete derivation. As the partition function does not depend on action $a_t^i$, we can ignore it in practice:
\begin{equation}
    \piN(a_t^i|g,h_t)\propto\piL(a_t^i|g,h_t)\exp{(\alpha \gVL(g,h_t,a_t^i,u_t^i))}.\label{equ:sol_practice}
\end{equation}
We simply take the action with maximum proportion $a_t\gets \argmax_{a^i_t} \piN(a_t^i|g,h_t)$. It is worth mentioning that if we let $\alpha=0$ in \cref{equ:sol_practice}, we recover the methods that directly use LLM's prior as a policy, and if we let $\alpha\to +\infty$, we recover the methods that incorporate planning and action evaluations.


There are two key advantages of using \cref{equ:sol_practice}. Firstly, it updates the action distribution of policy $\piL$ in the direction suggested by {critic} $\gVL$ in a gradient-free way, which achieves policy improvement with much lower computation burden compared to gradient-based methods, especially when the actor is realized by LLMs. Secondly, the action distribution of the new policy $\piN$ is a balanced integration of the policy's prior based on past information and the critic's posterior based on predicted future information.

\section{Experiments}\label{sec:exp}

In this section, we benchmark our method \name{} on three benchmarks that cover high-level action space (\alfworld{} \citep{shridhar2021alfworld}), low-level action space (\babyai{} \citep{chevalierboisvert2019babyai}) and potentially infinite action space (\webshop{} \citep{yao2022webshop}). We evaluate the effectiveness of \name{} by answering the following questions: (1) Can \name{} outperform other decision-making with LLM-based methods? (\cref{sec:exp_perf}) (2) How does each component of \name{} contribute to its performance? (\cref{sec:exp_ablation}) (3) How do different large language models influence performance? (\cref{sec:exp_perf} and \cref{sec:exp_ablation}) (4) Is our method computationally consuming? (\cref{sec:cost}) (5) Why is \name{} effective? (\cref{sec:stat-analysis}). The code of \name{} is publicly available on GitHub\footnote{\url{https://github.com/drdh/LAC}} and website\footnote{\url{https://sites.google.com/view/lang-ac}}.



\subsection{Experiment Setup}


We compare our method with various decision-making with LLMs baselines. Here we briefly introduce these methods, and for more details, please refer to \cref{appx:exp_baseline}.


\begin{figure*}[t]
\centering
\includegraphics[width=\linewidth]{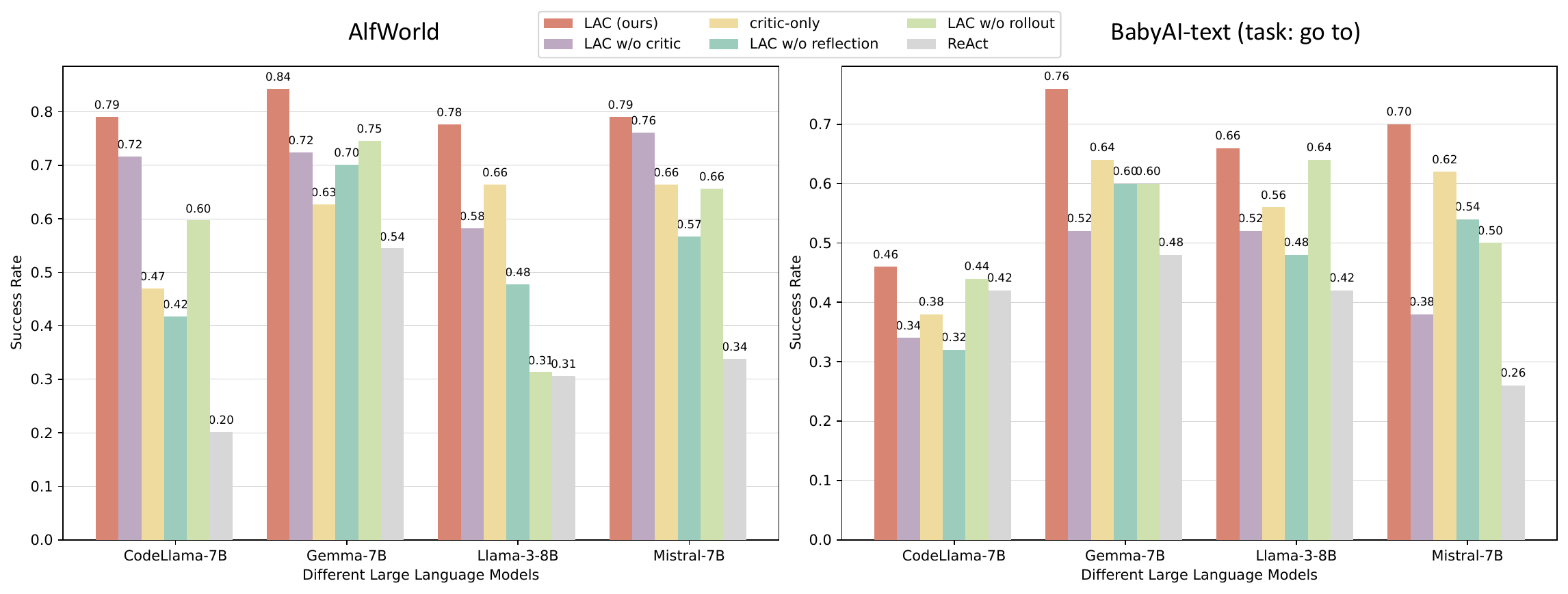}
\caption{Ablation studies in benchmarks \alfworld{} and \babyai{}.}\label{fig:ablation}
\end{figure*}

\textbf{Various baselines:} (1) \react{} \citep{yao2022react} combines reasoning and acting in the interaction with the environment and leverages the reasoning capabilities of LLMs to increase the probability of the LLM acting correctly as a policy. (2) \rap{} \citep{hao2023rap} utilizes LLMs as policy and world models and adopts tree-search planning methods to evaluate each possible action candidate. (3) \icpi{} \citep{brooks2024large} implements policy iteration using LLMs by predicting future trajectories and selecting the action with the highest predicted cumulative rewards. (4) \rafa{} \citep{liu2023rafa} evaluates each action candidate by tree-search and selects the action that may complete the most sub-goals. (5) \lats{} \citep{zhou2024lats} combines the reasoning, acting, and planning capabilities of LLMs with MCTS \citep{kocsis2006bandit} and external feedback mechanisms to enhance decision-making across various domains, achieving competitive results in web navigation.


We evaluate \name{} on three decision-making benchmarks with sparse rewards and distinct types of action space. 



\textbf{High-level actions: \alfworld{}} \citep{shridhar2021alfworld} is a widely used text-based household environment with 134 different evaluation tasks, which require the agent to achieve a goal through a sequence of high-level actions, \eg{} ``go to place-X'', ``take object-Y from place-X'', \etc{} The agent gets a reward of 1 if it achieves the goal, and 0 otherwise. The main challenge of this benchmark is to locate the target object and fulfill household work with commonsense knowledge of LLMs. Following \react{}, we evaluate all 134 unseen evaluation games in a task-specific setup.

\textbf{Low-level actions: \babyai{}} \citep{carta2023grounding} is a Grid World environment that extended from the BabyAI platform \citep{chevalierboisvert2019babyai}, in which the agent and objects are placed in a room of $8\times8$ tiles. The agent has 6 primitive actions: turn left, turn right, go forward, pick up, drop, toggle, to solve a task described in natural language (\eg{} ``Pick up the red box''). Similar to \alfworld{}, this benchmark also has binary rewards of 1 or 0. These tasks could be difficult because agents have to make a long-term plan, avoid obstacles, and find a short path to target objects based on partial observations that are described in natural language.

\textbf{Potentially infinite actions: \webshop{}} \citep{yao2022webshop} requires an agent to purchase a product based on instructions (e.g. ``I need a long clip-in hair extension which is natural looking, and price lower than 20.00 dollars'') through web interactions (e.g. search ``long clip-in hair extension'', click buttons such as ``[item ID]'' or ``back to search''). Within this context, the ``search'' and ``click'' actions can indeed lead to an unbounded set of potential actions, as the agent can continuously refine its queries and selections based on dynamic web results. Different from \alfworld{} and \babyai{}, the final reward in this benchmark is a continuous value between 0 and 1, depending on the degree to which the final purchased product meets the requirements.


To show the stability of \name{}, we adopt four open-source large language models from different organizations: \codellamaSeven{} \citep{roziere2023codellama}, \mistralSeven{} \citep{jiang2023mistral}, \gemmaSeven{} \citep{team2024gemma}, and \llamaEight{} \citep{meta2024llama3}.

\begin{figure*}[t]
\centering
\includegraphics[width=0.9\linewidth]{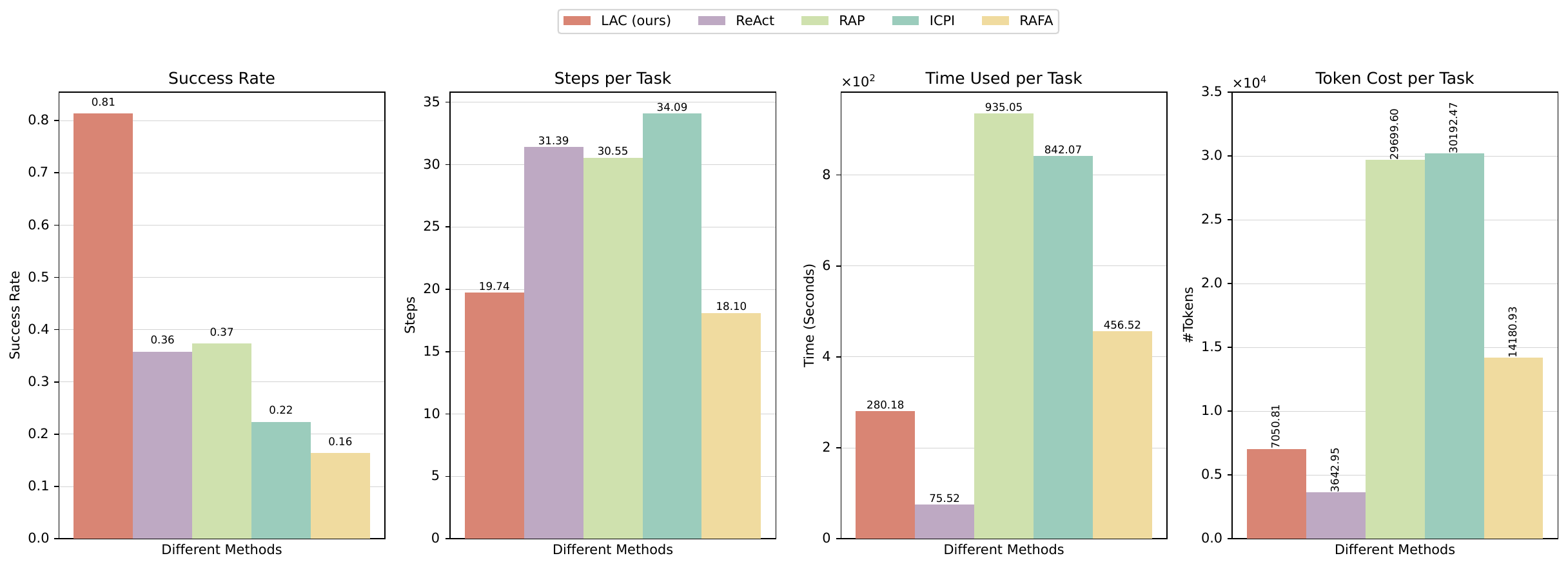}
\caption{Computational cost analysis of \name{} and baselines.}\label{fig:cost_main}
\end{figure*}

\subsection{Main Performance}\label{sec:exp_perf}

We report the results of our method \name{} compared with other baselines in \cref{fig:performance} (\alfworld{} and \babyai{}), \cref{fig:exp-webshop} (\webshop{}) and \cref{fig:performance-all} (more \babyai{} tasks). For all experiments, we set the temperature of LLMs to $0$, hence the generation is deterministic. For this reason, there is no error bar in the figure. 


\name{} outperforms all other baselines in both \alfworld{} and \babyai{}, and is even better than \gptFour+\react{} in most settings, which validates our method's effectiveness and stability. In \webshop{}, \name{} consistently outperforms other baselines, in terms of both accumulated reward and success rate across various base models. This further demonstrates the robustness of our method in handling more complex and open-ended action spaces.



\name{}'s superior performance stems from its effective integration of the decision-making information contained in the LLM prior with the decision-making insights derived from action evaluations that incorporate long-term planning. For better understanding, we have also provided illustrative examples for \alfworld{} and \babyai{} in \cref{fig:illustration} and \cref{fig:illustration-babyai} respectively. In summary, the policy and critic alone may make mistakes at different time steps, our \name{} can select the correct action through their integration. 

Regarding the performance of \name{} with different base models, we highlight two key findings: (1) Our method is general and can be adapted to various base models, and (2) stronger base models, such as \gemmaSeven{}, demonstrate higher performance when integrated with our approach. However, due to the incomplete public availability of training details for these base models, further in-depth analysis will require additional investigation.

Here we only compare \name{} with LLM-based methods, for more results regarding other baselines, \eg{}, finetuning the policy, LLM-based RL variants, and decision-transformer \citep{chen2021decision}, please refer to \cref{appx:more_baselines}.

\subsection{Ablation Studies}\label{sec:exp_ablation}

To investigate the contributions of each component of \name{}, we conduct elaborate ablation studies. There are several main components that characterize our method: (1) policy optimization step using action evaluations provided by critic; (2) action evaluations by extracting internal information of LLMs. Therefore, to show the contribution of each component, we design the following ablation studies: (1) \namewovaluecritic{} removes the policy optimization step from \name{}; (2) \nameworollout{} does not predict future trajectories during action evaluations; (3) \namewolangcritic{} removes the reflection procedure when sampling and evaluating candidate actions; (4) \nameonlyvaluecritic{} only uses critic's action evaluation information for action selection.



We report the results in \cref{fig:ablation}. \name{} is better than all other variants in both \alfworld{} and \babyai{}. Specifically, the performance decrease in \namewovaluecritic{} and \nameonlyvaluecritic{} compared with \name{} verify the effectiveness \name{}'s policy optimization with critic's action evaluations. And the result that \namewolangcritic{} and \nameworollout{} perform worse than \name{} also suggests the necessity for incorporating reflections and future trajectory predictions.







\subsection{Computational Cost Analysis}\label{sec:cost}

Our method conducts action evaluations by predicting future trajectories, which may bring extra computational cost per step. In \cref{fig:cost_main}, we compare computational costs concerning the number of tokens spent and running time between \name{} and other baselines. Specifically, though \name{} has a higher computational cost per step due to the extra inference procedure of critics and the forward model, the total cost of \name{} is still lower than most LLM-based baselines because \name{} requires fewer steps to finish each task. This is due to \name{}'s higher success rate, enabling it to complete tasks within the maximum step limit, while other baselines often reach this limit without completing the tasks. If we only consider successful tasks, the step cost is similar across methods: \name{}: $15.32$ steps, \react{}: $17.75$ steps, and \rap{}: $16.36$ steps. For more computational cost analysis of other baselines, please refer to \cref{appx:cost_more_baselines}.

\begin{table*}[ht]
\centering
\caption{Pearson correlations between Q-values and timesteps.}\label{tab:Pearson}
\begin{tabular}{@{}lcc@{}}
\toprule
                         & \multicolumn{1}{l}{Successful Trajectory (1843 steps in total)} & \multicolumn{1}{l}{Failed Trajectory (1092 steps in total)} \\ \midrule
$\log P$(``GOOD'')              & 0.35 ± 0.18                                                     & -0.37 ± 0.19                                                \\
$\log P$(``BAD'')               & -0.32 ± 0.19                                                    & 0.38 ± 0.19                                                 \\
Q-value of chosen action & 0.34 ± 0.18                                                     & -0.41 ± 0.18                                                \\ \bottomrule
\end{tabular}
\end{table*}

\begin{table*}[ht]
\centering
\caption{Confidence analysis of policy improvement.}\label{tab:confidence}
\begin{tabular}{@{}lccc@{}}
\toprule
Case (Proportion)                           & Prior Policy Conf. & Q-function Conf. & the Improved Policy Conf. \\ \midrule
Both agree (49.81\%)                        & 0.49 (\cmark)           & 0.21 (\cmark)         & 0.74                      \\
Prior agrees, Q disagrees (29.12\% cases)   & 0.34 (\cmark)           & 0.06 (\xmark)         & 0.28                      \\
Prior disagrees, Q agrees (19.28\% cases)   & 0.21 (\xmark)           & 0.34 (\cmark)         & 0.22                      \\
Prior disagrees, Q disagrees (1.79\% cases) & 0.06 (\xmark)           & 0.03 (\xmark)         & 0.07                      \\ \bottomrule
\end{tabular}
\end{table*}

\subsection{Statistical Analyses}\label{sec:stat-analysis}

Directly interpreting LLM-based decision-making is inherently challenging due to two factors: (1) LLMs are black-box models, making it difficult to directly explain their outputs; (2) In sequential decision-making tasks, ground-truth actions are typically unavailable for analyzing each action the policy selects—only task-level success or failure signals are observed.

Therefore, we provide statistical analyses to shed light on why the two key components of \name{}—Q-function estimation and gradient-free policy improvement—work effectively in practice.

\paragraph{Q-function estimation reflects task progression} 
We estimate Q-values using LLM-internal beliefs (log-probabilities of ``GOOD''/``BAD'' tokens). To validate that these Q-values meaningfully track task success, we compute the \textbf{Pearson correlation} between Q-values and timesteps within each trajectory. The intuition is: if an agent is on a successful path, Q-values should increase as it progresses (and decrease otherwise). The results in \cref{tab:Pearson}, averaged across 134 \alfworld{} tasks, confirm this trend: successful trajectories show increasing Q-values, while failed ones show the opposite. This demonstrates that our Q-function captures the evolving success likelihood of the policy throughout a trajectory.

\paragraph{Policy improvement reflects model confidence}
Our gradient-free policy improvement balances the prior policy and the Q-function based on \textbf{relative confidence}. Specifically, we define a model's confidence as the difference between its top two scores: (1) For the prior policy: $\log P(a_1|s) - \log P(a_2|s)$; (2) For the Q-function: $Q(a_1|s)-Q(a_2|s)$, where $a_1$ and $a_2$ are the actions with the highest and second-highest probability (or Q-value), respectively, under the corresponding model. 

We analyze which action (from the prior policy or Q-function) is selected by the improved policy and what confidence each model had in its choice. \cref{tab:confidence} summarizes the outcomes. This analysis shows that: (1)When both models agree, the confidence of the improved policy is highest; (2) When they disagree, the improved policy tends to trust the more confident model; (3) In low-confidence cases, the policy remains conservative. These results suggest our method implicitly aggregates decision knowledge from both sources by weighting based on model confidence, enabling effective policy improvement without explicit gradients.

\section{Discussion}\label{sec:discuss}

In this work, we introduce a novel LLM-based Actor-Critic algorithm \name{} that leverages action evaluations to optimize the prior policy of LLM, enabling an effective integration of the decision-making insights derived from LLM prior and action evaluations that incorporate long-term planning. Compared with previous methods, \name{} achieves high performance on three benchmarks that cover various action spaces even using lightweight open-source LLMs.

Despite the advancement, our method also has limitations and inspires possible future directions. Firstly, the reflection of \name{} is only used before action generation, which can also be applied after action generation. For example, it can provide reflections for predicted future trajectories to re-sample candidate actions if the previous candidate actions all fail to complete the target task. Secondly, though we only expand one node for each candidate action for simplicity and efficiency and find it works effectively, \name{} can also adopt tree-search to provide a more accurate assessment of candidate actions. Thirdly, though \name{} could deal with scenarios with continuous final reward empirically by only treating getting the highest reward as success, it is an exciting future direction to develop a more principal method to deal with such situations. Last but not least, the effectiveness of \name{} when applied to larger and more powerful LLM models needs further investigation.


\section*{Impact Statement}

Our method is built upon open-source large language models (LLMs). Like other methods that use LLMs, our method also inherits some benefits and challenges from LLMs. For the benefits, our method directly exploits the prior knowledge from LLMs, which may reduce potential carbon costs compared with training policies from scratch. For the challenges, our method might be susceptible to producing unintended output when confronted with harmful input, such as unethical text or input intended for adversarial attacks. To solve this problem, we suggest a thoughtful deployment of our method, such as adding a filtering component.




\bibliography{icml2025}
\bibliographystyle{icml2025}

\newpage
\appendix
\onecolumn

\section{Extra results}\label{appx:results}

\subsection{Results of other tasks in \babyai{}}\label{appx:performance-all}

\begin{figure*}[htp]
\centering
\includegraphics[width=\linewidth]{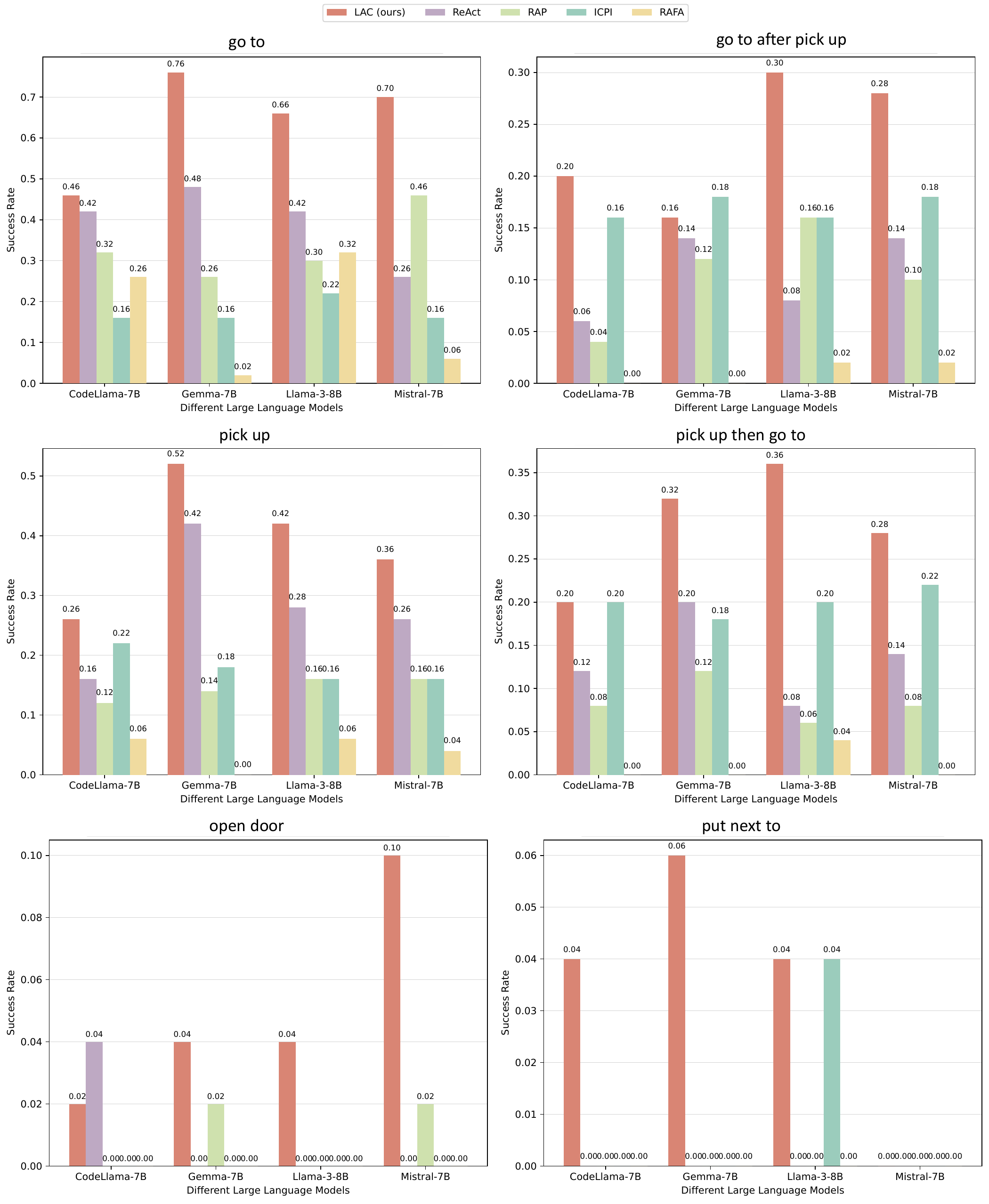}
\caption{Performance of our \name{} compared with various baselines in all tasks from \babyai{}.}\label{fig:performance-all}
\end{figure*}

For a complete comparison, we show the performance of \name{} and baselines in other tasks from \babyai{} in \cref{fig:performance-all}. Our \name{} outperforms all other baselines, which further validates the effectiveness of \name{}.





\subsection{Results of using other definition of \texorpdfstring{$\gVL$}{}}\label{appx:ablation-Q}

\begin{figure*}[ht]
\centering
\includegraphics[width=\linewidth]{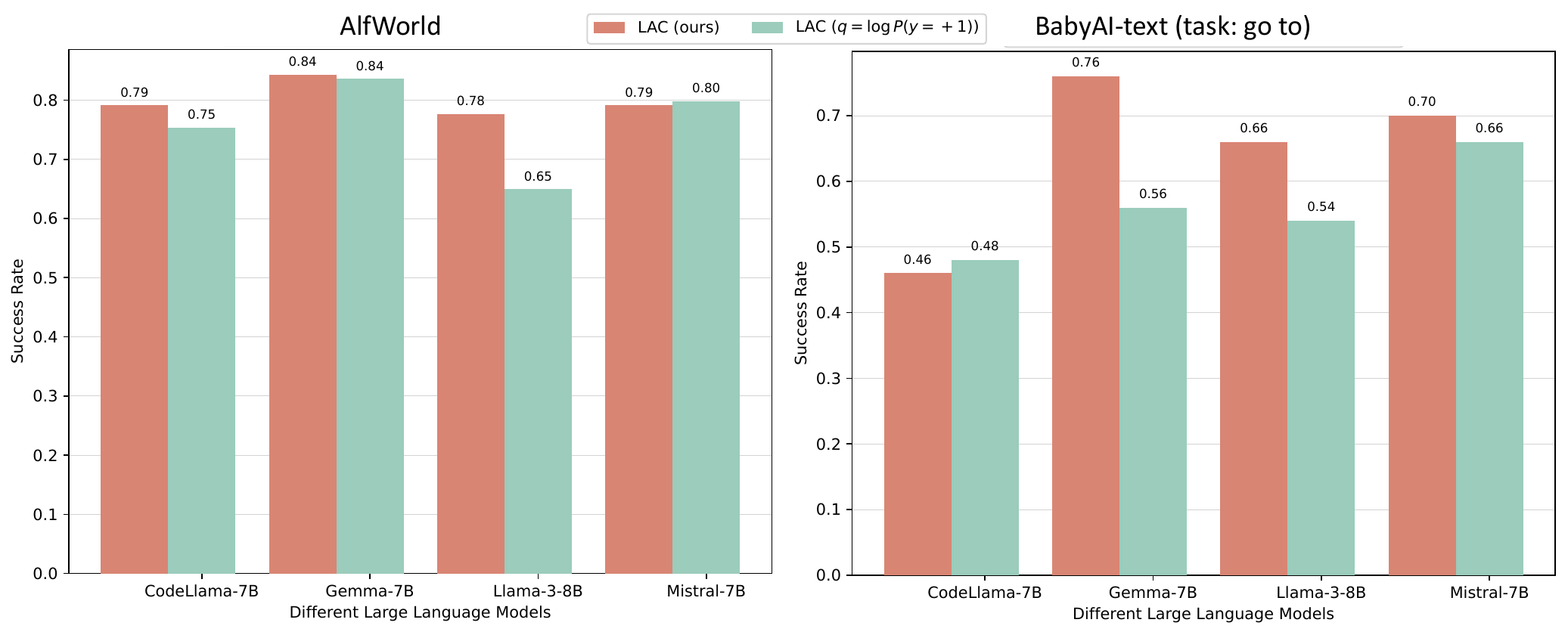}
\caption{Performance of \name{} when using different definition of critic $\gVL$}\label{fig:ablation-Q}
\end{figure*}

In \name{} we define critic $\gVL$ as $\gVL(g,h_t,a_t^i) = \log \frac{P(y_w|g,h_t,a_t^i)}{P(y_l|g,h_t,a_t^i)}$. There are also other definitions, for example, the simplest variant is $\gVL(g,h_t,a_t^i) = \log P(y_w|g,h_t,a_t^i)$. In this subsection, we provide a performance comparison between them in \cref{fig:ablation-Q}. \name{} outperforms the variant in most situations across tasks and models. We speculate that this is because \name{} uses more information, \ie{}, both $P(y_w|g,h_t,a_t^i)$ and $P(y_l|g,h_t,a_t^i)$, than the variant, and the evaluation might be more accurate and more stable. There might be other definitions of $\gVL$ and among them, our $\gVL$ is simple and effective.

\subsection{Results of directly prompting the LLMs to output action evaluation}\label{appx:direct_eval}

To further demonstrate the strengths of our critic $\gVL$. We conducted additional experiments comparing the performance of \name{} and \name{} w/ direct evaluation on the \webshop{} benchmark. \name{} generates value-based evaluations by extracting LLMs' internal beliefs of success and failure as described in \cref{equ:def_q}. For \name{} w/ direct evaluation, we prompt the LLMs to directly output the probability of success $P(y_w)$, while keeping all other components unchanged. The Q-value is then calculated as $\log\frac{P(y_w)}{1-P(y_w)}$. 
 
The results, presented in \cref{tab:direct_eval_succ} and \cref{tab:direct_eval_rew}, show that our \name{} method outperforms \name{} w/ direct evaluation in terms of success rate and reward across most base models. Analysis of the resulting trajectories revealed that \name{} w/ direct evaluation often produces a non-informative success probability (\eg{}, $P(y_w)=0.5$), leading to ineffective improvements in policy.

\begin{table}[htbp]
\centering
\caption{Success rate of \name{} and \name{} w/ direct evaluation in \webshop{}}\label{tab:direct_eval_succ}
\begin{tabular}{@{}lllll@{}}
\toprule
                       & \codellamaSeven{} & \gemmaSeven{}    & \llamaEight{} & \mistralSeven{}  \\ \midrule
\name{}                & \textbf{32\%}  & \textbf{46\%} & \textbf{39\%}       & \textbf{32\%} \\
\name{} w/ direct evaluation & 27\%           & 29\%          & 33\%       & 24\%          \\ \bottomrule
\end{tabular}
\end{table}

\begin{table}[htbp]
\centering
\caption{Rewards of \name{} and \name{} w/ direct evaluation in \webshop{}}\label{tab:direct_eval_rew}
\begin{tabular}{@{}lllll@{}}
\toprule
                    & \codellamaSeven{} & \gemmaSeven{}    & \llamaEight{} & \mistralSeven{}    \\ \midrule
\name{}                & \textbf{0.5840} & \textbf{0.7237} & \textbf{0.6733} & 0.6299          \\
\name{} w/ direct evaluation & 0.5636          & 0.6975          & 0.6453          & \textbf{0.6333} \\ \bottomrule
\end{tabular}
\end{table}

\subsection{Comparison of our reflection component and chain-of-thought}\label{appx:reflection_cot}

In this subsection, we empirically compare the reflection method used in \name{} and the Chain-of-Thought (CoT) methods \cite{wei2022cot, kojima2022zerocot}.

Though the reflection draws inspiration from CoT, it is not trivial to determine what the reflection content should be. Our method generates judgments based on previous actions and their outcomes, whereas CoT uses arbitrary thought generation without a structured focus on past mistakes. The reflection is specially designed for decision-making tasks and owns two advantages over naive CoT:

\paragraph{Strength 1.} Our reflection can improve the policy directly by avoiding previous mistakes when the policy is conditioned on its generation.

To substantiate this claim, we conducted experiments comparing the performance of prior policy w/ reflection and piror policy w/ CoT on the \webshop{} benchmark. Here are the details of these two variants:

(1) prior policy w/ reflection: We remove all components of \name{}, leaving the policy and the reflection unchanged. Specifically, at each step, after observing the action results, the reflection first generates some judgments on previous actions, and then the policy selects the next action based on the judgments. (2) prior policy w/ CoT: We remove all components of \name{}, except the policy. Additionally, we equip the policy with CoT by adding ``Let's think step by step'' to the prompt. Specifically, at each step, before choosing the next action, the CoT prompting component first outputs arbitrary thoughts that may help solve the task.

As shown in \cref{tab:reflection_cot_prior_succ} and \cref{tab:reflection_cot_prior_rew}, prior policy w/ reflection consistently surpasses prior policy w/ CoT across most base models in terms of both success rate and reward. By analyzing the results, we found that prior policy w/ CoT may make the same mistake multiple times and get stuck at this mistake, while prior policy w/ reflection can largely avoid seen mistakes.

\begin{table}[htbp]
\centering
\caption{Success rate of prior policy w/ reflection and prior policy w/ CoT in \webshop{}}\label{tab:reflection_cot_prior_succ}
\begin{tabular}{@{}lllll@{}}
\toprule
                           & \codellamaSeven{} & \gemmaSeven{}    & \llamaEight{}    & \mistralSeven{}  \\ \midrule
prior policy w/ reflection & \textbf{21\%}  & \textbf{46\%} & \textbf{39\%} & \textbf{31\%} \\
prior policy w/ CoT        & \textbf{21\%}  & 20\%          & 31\%          & {20\%} \\ \bottomrule
\end{tabular}
\end{table}

\begin{table}[htbp]
\centering
\caption{Rewards of prior policy w/ reflection and prior policy w/ CoT in \webshop{}}\label{tab:reflection_cot_prior_rew}
\begin{tabular}{@{}lllll@{}}
\toprule
                    & \codellamaSeven{}  & \gemmaSeven{}      & \llamaEight{}      & \mistralSeven{}    \\ \midrule
prior policy w/ reflection & \textbf{0.5739} & \textbf{0.6564} & \textbf{0.6556} & \textbf{0.6288} \\
prior policy w/ CoT        & 0.5520          & 0.5347          & 0.6379          & 0.4671          \\ \bottomrule
\end{tabular}
\end{table}

\paragraph{Strength 2.} Our reflection can be seamlessly integrated with critic $\gVL$, which helps the critic to generate more accurate value-based evaluations.

To further evaluate this integration, we compare the performance of \name{} and \name{} w/ CoT on the \webshop{} benchmark. The details of the two methods are as follows:

(1) \name{}: Our original method. (2) \name{} w/ CoT: We replace the reflection component of \name{} with CoT and keep other components unchanged.

We show the results in \cref{tab:reflection_cot_lac_succ}  and \cref{tab:reflection_cot_lac_rew}. Our method \name{} consistently outperforms \name{} w/ CoT regarding success rate and reward across all evaluated base models. This is because, without reflection's judgment on previous steps, the critic may output inaccurate value-based estimations, hindering the policy improvement phase.

\begin{table}[htbp]
\centering
\caption{Success rate of \name{} and \name{} w/ CoT in \webshop{}}\label{tab:reflection_cot_lac_succ}
\begin{tabular}{@{}lllll@{}}
\toprule
    & \codellamaSeven{} & \gemmaSeven{}    & \llamaEight{}    & \mistralSeven{}  \\ \midrule
\name{}        & \textbf{32\%}  & \textbf{46\%} & \textbf{39\%} & \textbf{32\%} \\
\name{} w/ CoT & 27\%           & 29\%          & 33\%          & 24\%          \\ \bottomrule
\end{tabular}
\end{table}

\begin{table}[htbp]
\centering
\caption{Rewards of \name{} and \name{} w/ CoT in \webshop{}}\label{tab:reflection_cot_lac_rew}
\begin{tabular}{@{}lllll@{}}
\toprule
     & \codellamaSeven{}  & \gemmaSeven{}      & \llamaEight{}      & \mistralSeven{}    \\ \midrule
\name{}        & \textbf{0.5840} & \textbf{0.7237} & \textbf{0.6733} & \textbf{0.6299} \\
\name{} w/ CoT & 0.5734          & 0.5270          & 0.6515          & 0.5101          \\ \bottomrule
\end{tabular}
\end{table}

In summary, the reflection's focused evaluations of past actions provide critical advantages over the simple reflections of CoT, leading to improved performance in decision-making tasks.

\subsection{Comparison of \name{} with more baselines on \alfworld{}}\label{appx:more_baselines}

\begin{figure*}[ht]
\centering
\includegraphics[width=\linewidth]{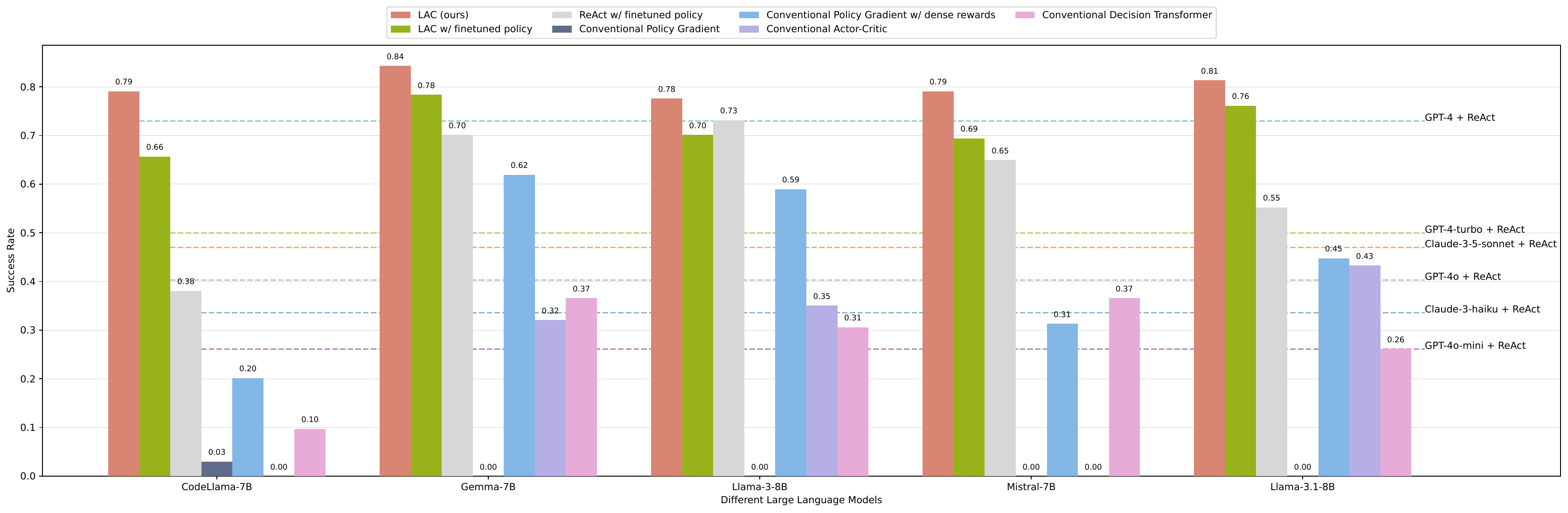}
\caption{Performance of our \name{} and \name{}'s variants compared with various baselines in benchmark \alfworld{}.}
\label{fig:performance-more_baselines}
\end{figure*}

In this subsection, we compare \name{} with more baselines including some traditional RL methods implemented using LLMs on \alfworld{} \citep{shridhar2021alfworld}. The comparison is shown in \cref{fig:performance-more_baselines}.

\subsubsection{Finetune policy}

While in \name{} we fine-tune the model for generating reflections using a few trajectories, it is also possible to fine-tune the policy to generate actions using those trajectories. Therefore, a potential baseline could be fine-tuning the policy in \react{}. To demonstrate the improvement brought by fine-tuning the policy, we fine-tune the policy in \react{} \citep{yao2022react} and show the results in \cref{fig:performance-more_baselines}. We also show the results of \name{} w/ fine-tuned policy in \cref{fig:performance-more_baselines}. In brief, \react{} w/ fine-tuned policy is a strong baseline compared with other baselines, but still inferior to our method \name{} and \name{} w/ fine-tuned policy. Compared to \name{}, the underperformance of \name{} w/ fine-tuned policy arises from its tendency to overfit the training trajectories. This overfitting causes the policy to favor actions that are more frequent in the dataset, potentially leading to suboptimal action selection. 

For example, in the \alfworld{} training dataset, the action ``take an apple from X'' occurs frequently. After fine-tuning, the policy may disproportionately generate this action, even when it is irrelevant to the current goal. One case is that the current goal is to ``heat some egg and put it in the garbage can''. When the agent sees an ``apple 2'' in ``fridge 1'', it generates and selects an irrelevant action ``take apple 2 from fridge 1'', which does not align with the task.

This tendency towards overfitting arises because the complexity of the policy function, which maps states $s$ to actions $a$, often exceeds that of the critic. The policy often has to capture a wide variety of potential actions for each state, particularly in complex environments. However, the quite limited training dataset in our setting restricts its ability to generalize effectively, resulting in memorization of specific actions rather than flexible decision-making. In contrast, the model for reflection generation and rollouts focuses on capturing more predictable dynamics of the environment and simpler evaluation criteria. This typically requires simpler mappings than those needed for the policy, thus avoiding overfitting.


\subsubsection{LLM-based RL variants}

We also include some LLM-based RL variants as baselines to show the superiority of \name{} over conventional RL algorithms. We design three LLM-based RL variants that are built upon pre-trained LLMs and directly extract actions/values information from LLMs without adding action/value heads, namely Conventional Policy Gradient, Conventional Policy Gradient w/ dense rewards and Conventional Actor-Critic in \cref{fig:performance-more_baselines}.

For the implementation of the Conventional Policy Gradient, we need the probability of actions and the returns. To obtain the probability of actions, we directly use LLM to compute the conditional probability of each token in action $a_{i}=[w_{1},w_{2},\cdots,w_{|a_{t}|}]$ given the goal $g$, history $h_{t}$ and then calculate their product: $$\pi(a_{t}|g,h_{t})=\Pi_{j=1}^{|a_{t}|}P_{LLM}(w_{j}|g,h_{t},w_{<j})$$ in which $P_{LLM}(w_{j}|g,h_{t},w_{<j})$ is the probability of token $w_{j}$ given goal $g$, history $h_{t}$ and previous tokens $w_{<j}$ computed by LLM. Then we regard the cumulative future rewards as the return $G_{t}$, which is $+1$ for successful trajectories and $-1$ for failed trajectories in the tasks we considered. Finally, the gradient of policy is $\mathbb{E}[\sum_{t}\nabla{\rm log}\pi(a_{t}|g,h_{t})G_{t}]$. Conventional Policy Gradient w/ dense rewards is similar to Conventional Policy Gradient except that we manually add intermediate rewards for each step, and then use the cumulative future rewards as the return $G_{t}$.

For the implementation of the Conventional Actor-Critic, we additionally need a critic to estimate action values. As it is possible to train a new value head using only 18 trajectories, we instead approximate the action value similar to $\gVL$ in our method \name{}, \ie{} $$\gVL(g,h_{t},a_{t},u_t)={\rm log}\frac{P_{LLM}(y_w|g,h_{t},a_{t},u_t)}{P_{LLM}(y_l|g,h_{t},a_{t},u_t)}$$ in which $P_{LLM}(y_{\{w,l\}}|g,h_{t},a_{t},u_t)$ is the output probability of special positive/negative tokens like GOOD or BAD that indicate positive/negative results as LLM's belief on success/failure. Finally, the gradient of policy is $\mathbb{E}[\sum_{t}\nabla{\rm log}\pi(a_{t}|g,h_{t})\gVL(g,h_{t},a_{t},u_t)]$.

In summary, Conventional Policy Gradient exhibits almost all zero performance, which is due to the extremely sparse reward problems, compared with Conventional Policy Gradient w/ dense rewards. Conventional Actor-Critic demonstrates non-zero performance only on some stronger LLMs like \gemmaSeven{} \citep{team2024gemma}, \llamaEight{} \citep{meta2024llama3} and \llamaThreeOneEight{} \citep{meta2024llama31}, which may be because the optimization method of conventional actor-critic is not suitable in insufficient data settings.

\subsubsection{Decision Transformer}

In addition to the aforementioned LLM-based RL variant, Decision Transformer \citep{chen2021decision} is also a potential solution in combining RL and transformer-based LLMs. We fine-tune pretrained LLMs in a similar way as conventional decision transformers. We construct a dataset using decision-transformers' trajectory representation: $\tau=[R_{1},s_{1},a_{1},R_{2},s_{2},a_{2},\cdots]$, in which $R_{t}$ is return-to-go, \ie{}, +1 for successful trajectories and -1 for failed trajectories in our extremely sparse reward settings. Then we fine-tune LLMs with next-token prediction loss on these trajectories. During execution, we insert +1 before state $s_{t}$ to specify the desired outcome. The results are shown in \cref{fig:performance-more_baselines} as Conventional Decision Transformer. In short, Conventional Decision Transformer exhibits a similar performance to \react{}, which may be because the 18 trajectories are insufficient for fine-tuning decision transformers.

Our method \name{} is better than all considered baselines because of its ability to handle extremely sparse reward problems using LLM's prior knowledge and to fully utilize insufficient data.






\subsection{Computational cost analysis of \name{} with more baselines in \alfworld{}}\label{appx:cost_more_baselines}

\begin{figure*}[ht]
\centering
\includegraphics[width=\linewidth]{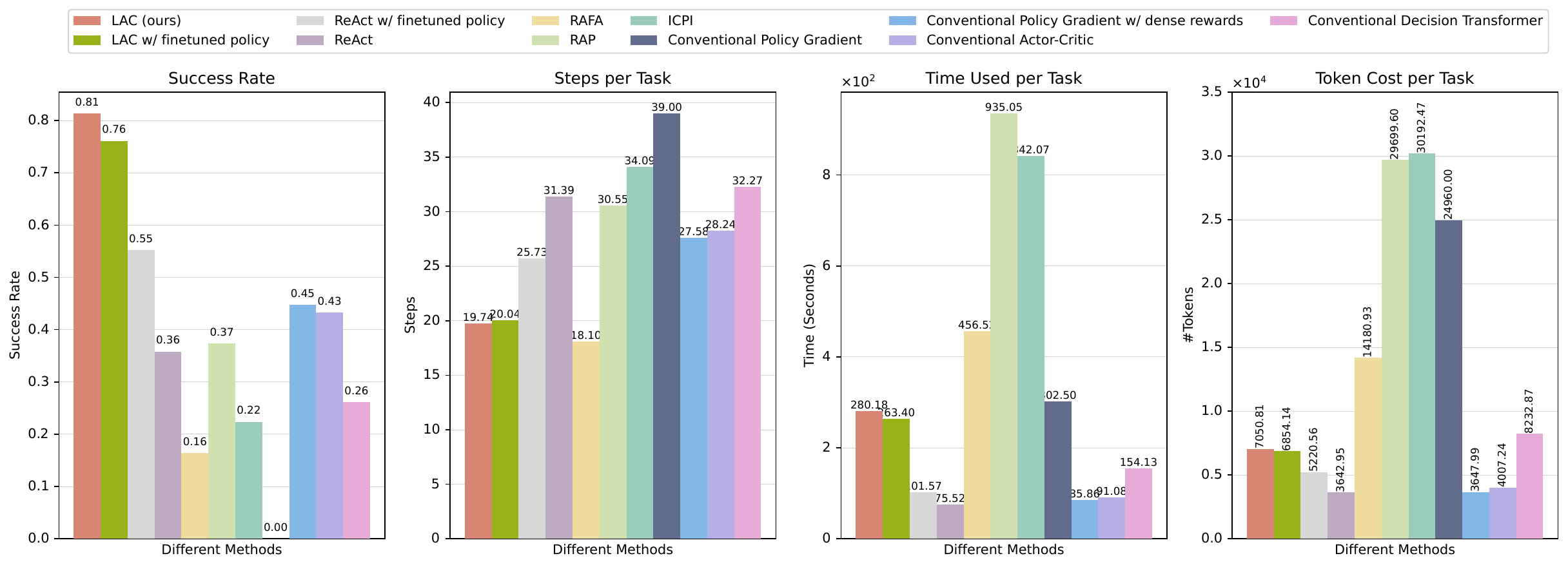}
\caption{Computational cost analysis of our \name{} compared with 
various baselines in benchmarks \alfworld{}. Though LAC may have a higher computational cost per step due to the extra inference procedure of critics and the forward model, the total cost of LAC is still lower than most LLM-based baselines because LAC requires fewer steps to finish each task.}
\label{fig:computation-cost}
\end{figure*}

In this subsection, we demonstrate the computation cost of \name{} and other baselines in \cref{fig:computation-cost}. We show the success rate, steps per task, time used per task, and token cost per task respectively. Specifically, though \name{} has a higher computational cost per step due to the extra inference procedure of critics and the forward model, the total cost is still lower than most LLM-based baselines because \name{} has a higher success rate and requires fewer steps to finish each task.

\subsection{Illustration of \alfworld{} and \babyai{}}

\begin{figure*}[htp]
\centering
\includegraphics[width=\linewidth]{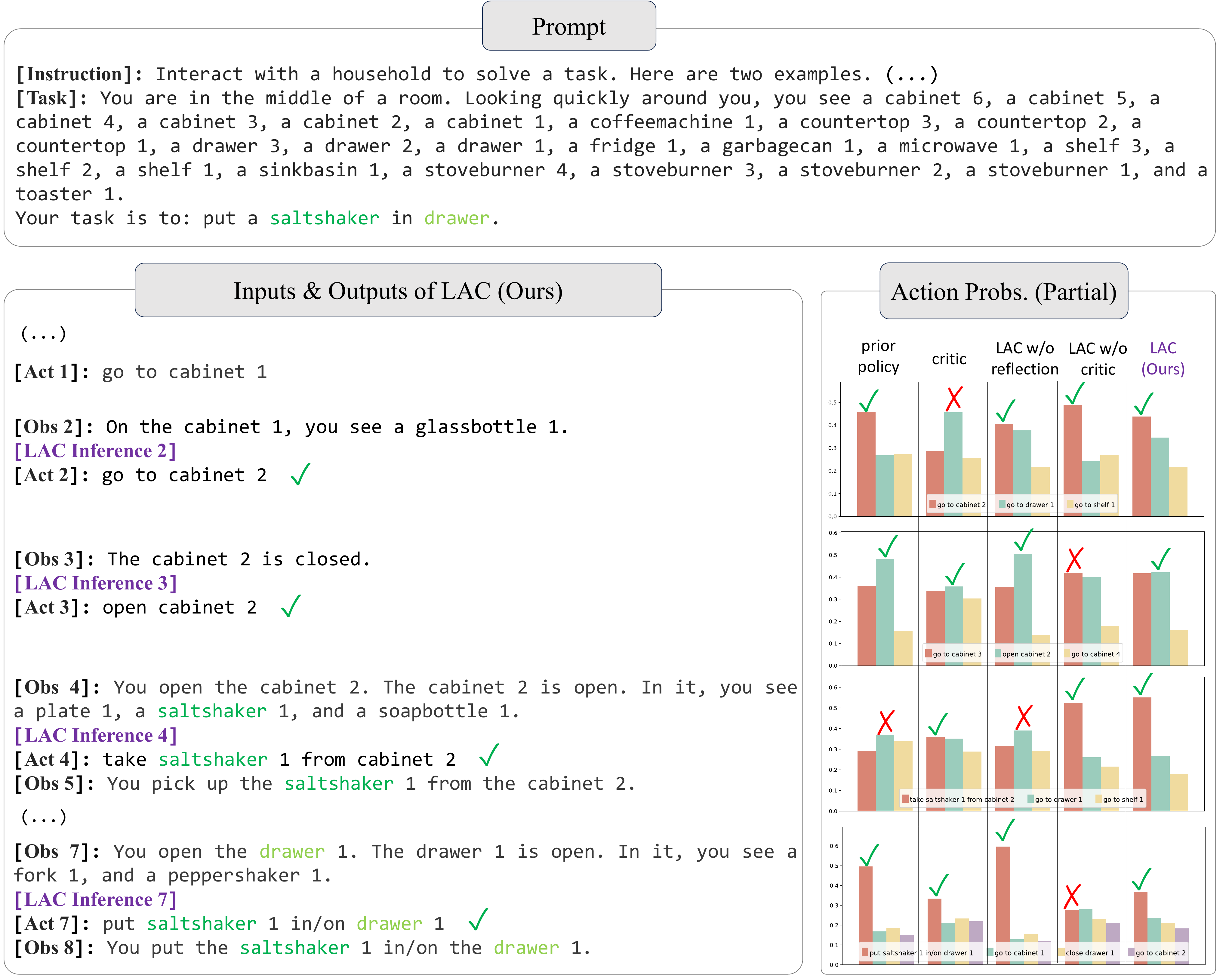}
\caption{An illustrative explanation of our method \name{} in \alfworld{}. The histogram on the right shows the action probabilities of different methods. While LLM's prior policy ($\piL$) and critic, as well as \name{} w/o reflection, make mistakes at different time steps, \name{} (ours) can select the correct action by optimizing the policy given action evaluation. The \name{} inference step is detailed in \cref{fig:framework}.
}
\label{fig:illustration}
\end{figure*}

\begin{figure*}[htp]
\centering
\includegraphics[width=\linewidth]{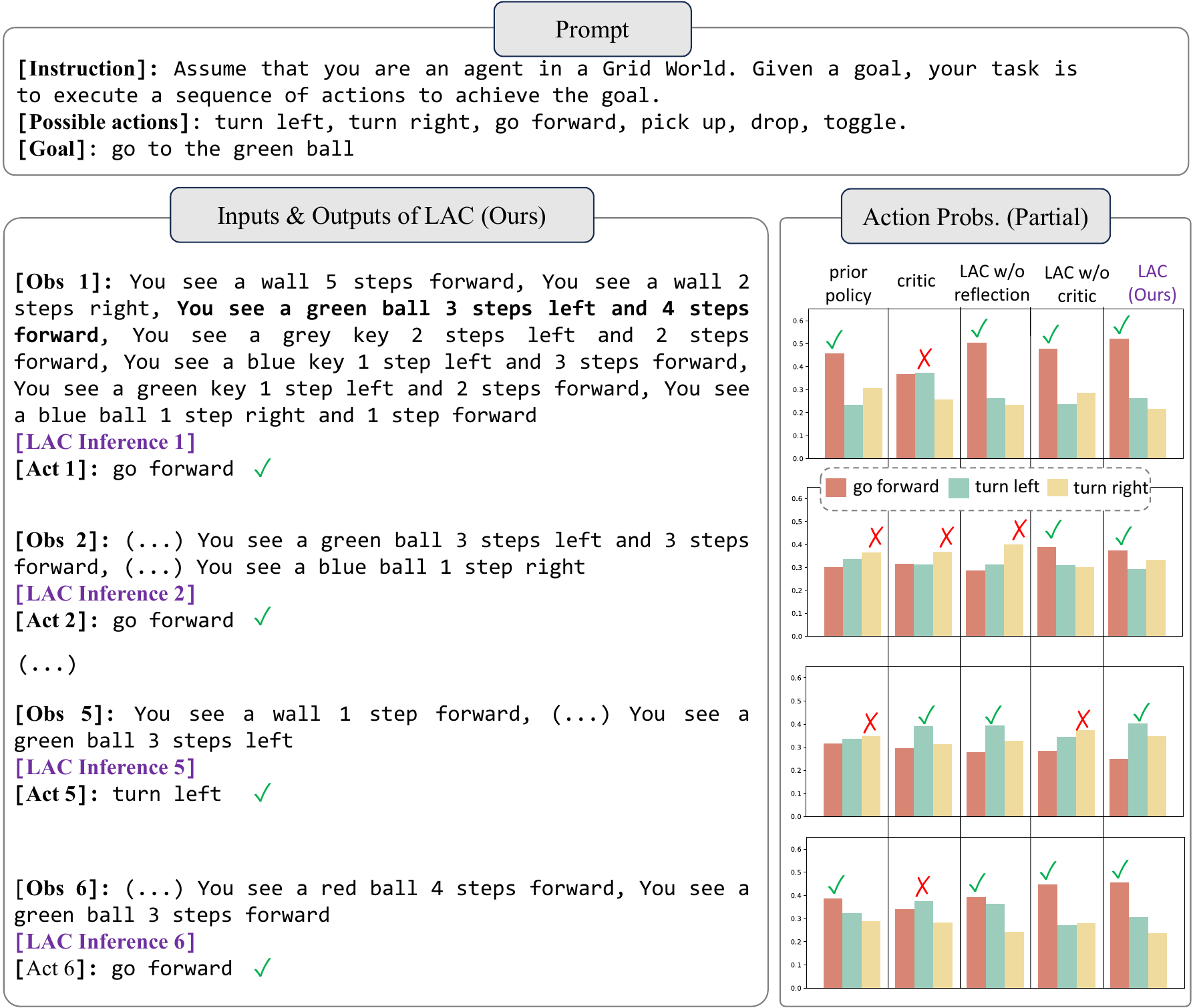}
\caption{An illustrative explanation of our method \name{} in \babyai{}. The histogram on the right shows the action probabilities of different methods. While LLM's prior policy ($\piL$) and critic, as well as \name{} w/o reflection, make mistakes at different time steps, \name{} (ours) can select the correct action by optimizing the policy given action evaluation. Please refer to \cref{tab:ablation_example} for the full trajectory.}
\label{fig:illustration-babyai}
\end{figure*}

We show the illustrative example of \alfworld{} and \babyai{} in \cref{fig:illustration} and \cref{fig:illustration-babyai}, respectively. 

\cref{fig:illustration} presents a concrete scenario where the agent's goal is to ``put a saltshaker in the drawer''. At a critical decision step (Step 4), we observe the following intuitive distinction between components: 
(1) LLM-based prior policy alone mistakenly suggests ``go to drawer 1'' because the base LLM model overlooks that the agent has already found the correct object (``saltshaker 1'') in cabinet 2. This error exemplifies the common hallucination problem in LLMs, which occurs when the model disregards previous states and incorrectly recommends irrelevant actions.
(2) In contrast, the critic suggests ``take saltshaker 1 from cabinet 2'' because it evaluates potential actions by predicting future trajectories and determines that this action will successfully pick up the correct object.
(3) Our method leverages these distinct insights by optimizing the prior policy's action distribution based on the critic's evaluation. It effectively corrects the errors introduced by the prior policy, balancing the strengths of prior policy (flexible but sometimes inaccurate) and critic evaluations (accurate but computationally intensive).


This analysis shows that: (1)When both models agree, the confidence of the improved policy is highest; (2) When they disagree, the improved policy tends to trust the more confident model; (3) In low-confidence cases, the policy remains conservative.

These results suggest our method implicitly aggregates decision knowledge from both sources by weighting based on model confidence, enabling effective policy improvement without explicit gradients.

\subsection{Results of different critic improvement methods}\label{appx:finetune_other}

Empirically, we found that the critic can be improved via fine-tuning on a few trajectories. Please refer to \cref{appx:critic_improve} for more fine-tuning details. To show the effectiveness of fine-tuning, we present the performance of \name{} and other variants when we just add these examples into the prompt, \ie{}, in-context learning, on task ``go to'' and ``pick up'' from \babyai{} in \cref{tab:finetune} and \cref{tab:finetune_other} respectively. We also show the the performance improvement if we do fine-tuning in the parentheses.
This result indicates that fine-tuning can incorporate extra knowledge into LLMs better than in-context learning in our case and improve the quality of action evaluation, yielding better performance. It is worth mentioning that our \name{} still outperforms baselines without fine-tuning.


\begin{table}[ht]
\vspace{-1em}
\caption{Performance of two critic improvement methods: in-context learning or fine-tuning.}\label{tab:finetune}
\centering
\begin{tabular}{@{}lllll@{}}
\toprule
                    & \codellamaSeven{}    & \gemmaSeven{}             & \llamaEight{}          & \mistralSeven{}        \\ \midrule
\name{}                & 0.30 ($\uparrow$ 0.16) & 0.62 ($\uparrow$ 0.14) & 0.32 ($\uparrow$ 0.34) & 0.24 ($\uparrow$ 0.46) \\
\namewolangcritic{}    & 0.30 ($\uparrow$ 0.02) & 0.58 ($\uparrow$ 0.02) & 0.38 ($\uparrow$ 0.10) & 0.26 ($\uparrow$ 0.28) \\
\namewovaluecritic{}   & 0.28 ($\uparrow$ 0.06) & 0.48 ($\uparrow$ 0.04) & 0.42 ($\uparrow$ 0.10) & 0.10 ($\uparrow$ 0.28) \\
\nameonlyvaluecritic{} & 0.42 ($\uparrow$ 0.04) & 0.40 ($\uparrow$ 0.24) & 0.34 ($\uparrow$ 0.22) & 0.38 ($\uparrow$ 0.24) \\ \bottomrule
\end{tabular}
\end{table}

\begin{table}[ht]
\centering
\caption{Performance of two critic improvement methods: in-context learning or fine-tuning.}\label{tab:finetune_other}
\begin{tabular}{lllll}
\toprule
                    & \codellamaSeven{}         & \gemmaSeven{}             & \llamaEight{}             & \mistralSeven{}           \\ \midrule
\name{}                & 0.20 ($\uparrow$ 0.06) & 0.22 ($\uparrow$ 0.20) & 0.34 ($\uparrow$ 0.08) & 0.20 ($\uparrow$ 0.16) \\
\namewolangcritic{}    & 0.16 ($\uparrow$ 0.08) & 0.32 ($\uparrow$ 0.04) & 0.32 ($\uparrow$ 0.04) & 0.22 ($\uparrow$ 0.06) \\
\namewovaluecritic{}   & 0.12 ($\uparrow$ 0.14) & 0.36 ($\uparrow$ 0.20) & 0.28 ($\uparrow$ 0.06) & 0.26 ($\uparrow$ 0.04) \\
\nameonlyvaluecritic{} & 0.22 ($\uparrow$ 0.04) & 0.24 ($\uparrow$ 0.26) & 0.16 ($\uparrow$ 0.16) & 0.16 ($\uparrow$ 0.26) \\ \bottomrule
\end{tabular}
\end{table}




\subsection{Analysis of the fine-tuning process in \name{}}\label{appx:finetuning-analysis}

\begin{figure*}[ht]
\centering
\includegraphics[width=\linewidth]{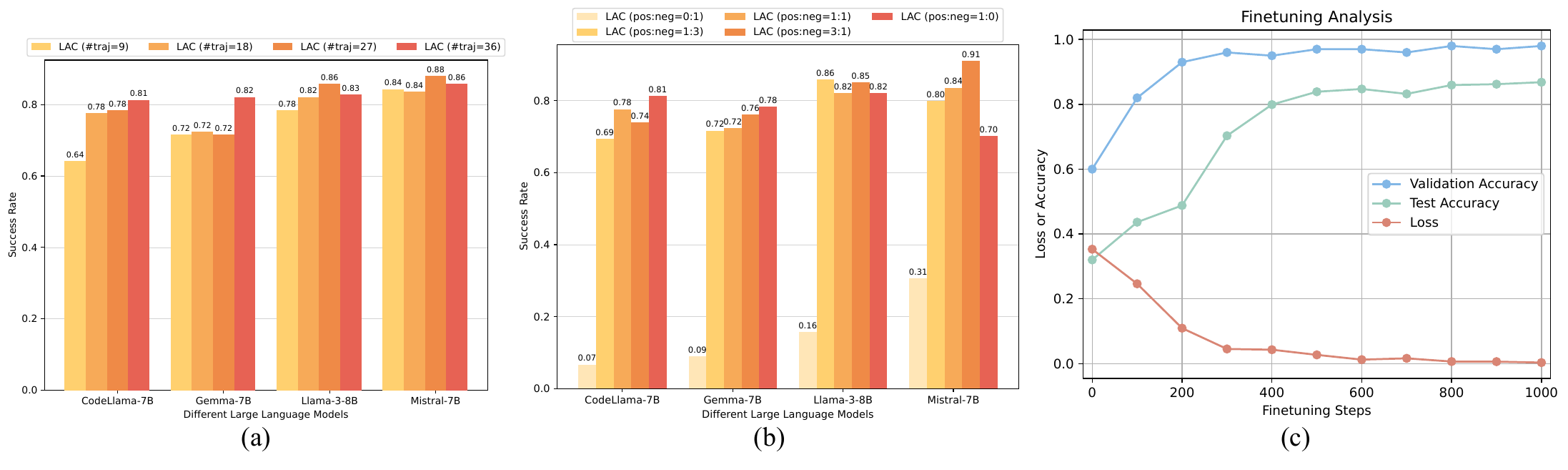}
\caption{Analysis regarding the fine-tuning process of our \name{}. (a) Influence of the fine-tuning data size. Larger data sizes (27, 36 trajectories) generally bring higher performance, but small data sizes (18 and even 9 trajectories) are already enough for our method to achieve outperformance. (b) Influence of the positive/negative data ratio. \name{} is robust to reasonable positive/negative ratios (1:3, 1:1, 3:1) while CodeLlama-7B and Gemma-7B-based \name{} even perform better given all positive data (1:0). (c) Learning curves of next-token prediction loss and positive/negative tokens prediction accuracy for CodeLlama-7B and \alfworld{}.}
\label{fig:finetuning-analysis}
\end{figure*}

In order to improve the quality of the reflections generated by LLM, we finetune the LLM that generates the reflections. In this section, we analyze the finetuning process, showing the impact of finetuning, as well as the impact of different data amounts and positive and negative sample ratios on task success rates. The comparison can be seen in \cref{fig:finetuning-analysis}.

In \cref{fig:finetuning-analysis} (a), we show the influence of fine-tuning data size. We use 9, 18, 27 and 36 trajectories to fine-tune LLMs, and show the final success rate on 134 evaluation tasks. In summary, larger data sizes (27 or 36 trajectories) generally bring higher success rate, while small data sizes (18 and even 9 trajectories in some cases) are already enough for \name{} to achieve outperformance.

\cref{fig:finetuning-analysis} (b) shows the influence of different positive/negative sample ratio (positive:negative = 0:1, 1:3, 1:1, 3:1 and 1:0) on final performance. We keep the total number of samples the same and just change positive/negative ratio. In short, our \name{} is robust to reasonable positive/negative ratios (\eg{} 1:3, 1:1, 3:1), while \name{} based on \codellamaSeven{} \citep{roziere2023codellama} and \gemmaSeven{} \citep{team2024gemma} even perform better when given all positive samples (1:0).

\cref{fig:finetuning-analysis} (c) shows the learning curves of the fine-tuning process. We plot the next prediction loss and positive/negative tokens prediction accuracy for \codellamaSeven{} \citep{roziere2023codellama}. In short, as the next token prediction loss decreases during fine-tuning, the accuracy of predicting the special tokens (GOOD or BAD) increases, which exhibits the effect of the fine-tuning process.

\subsection{Ablation studies on the impact of finetuning}

The finetuning process is important in our method, but it can also be replaced by in-context learning, where several examples are inserted in each input. Here we compare our method with \react{} regarding the effect of finetuning in \alfworld{}. The results, presented in \cref{tab:finetuning}, show that our method outperforms the baseline both with and without finetuning, and the performance can be further improved with finetuning. With finetuing, our method provides more consistent performance with different LLMs.

\begin{table}[ht]
\centering
\caption{Ablation studies on the impact of finetuning.}\label{tab:finetuning}
\begin{tabular}{@{}lcccc@{}}
\toprule
{ }                 & \multicolumn{1}{l}{\codellamaSeven{}} & \multicolumn{1}{l}{\gemmaSeven{}} & \multicolumn{1}{l}{\llamaEight{}} & \multicolumn{1}{l}{\mistralSeven{}} \\ \midrule
\name{} (\textit{w/o} finetuning)  & \textbf{0.39}                               & \textbf{0.59}                           & \textbf{0.71}                           & \textbf{0.57}                             \\
\react{} (\textit{w/o} finetuning) & 0.20                               & 0.54                           & 0.31                           & 0.34                             \\ \hline
\name{} (\textit{w/} finetuning)   & \textbf{0.79}                               & \textbf{0.84}                           & \textbf{0.78}                           & \textbf{0.79}                             \\
\react{} (\textit{w/} finetuning)   & 0.38                               & 0.70                           & 0.73                           & 0.65                             \\ \bottomrule
\end{tabular}
\end{table}

\subsection{Results of different hyper-parameter \texorpdfstring{$\alpha$}{}}\label{appx:alpha}

\begin{table}[ht]
\centering
\caption{Results of different hyper-parameter $\alpha$}\label{tab:alpha}
\begin{tabular}{@{}lllll@{}}
\toprule
                    & \codellamaSeven{} & \gemmaSeven{} & \llamaEight{} & \mistralSeven{} \\ \midrule
\name{} ($\alpha=1/2$) & \textbf{0.46}  & 0.54          & 0.62          & 0.68          \\
\name{} ($\alpha=1$)   & \textbf{0.46}  & 0.62          & 0.64          & 0.58          \\
\name{} ($\alpha=2$)   & 0.44           & \textbf{0.76} & \textbf{0.66} & 0.64          \\
\name{} ($\alpha=5$)   & \textbf{0.46}  & 0.72          & 0.62          & 0.64          \\
\name{} ($\alpha=10$)  & 0.40           & 0.58          & 0.60          & \textbf{0.70} \\ \bottomrule
\end{tabular}
\end{table}

The hyper-parameter $\alpha$ in \cref{equ:optimization} controls the deviation from the original policy $\piL$. In this subsection, we grid-search this hyper-parameter over $\{1/2,1,2,5,10\}$ in task ``go to'' of \babyai{}, then we fix $\alpha$ for other tasks: $\alpha=1$ for model \codellamaSeven{}, $\alpha=2$ for model \gemmaSeven{}, $\alpha=2$ for model \llamaEight{} and $\alpha=10$ for model \mistralSeven{}.

As for benchmark \alfworld{}, we fixed $\alpha=1$ in all experiments.

\subsection{Performance comparison in other benchmarks}
To further show the strength of our method, we conduct preliminary experiments on \crafter{} \cite{hafner2021crafter}, using the implementation from \balrog{} \cite{paglieri2024balrog}. \crafter{}  is a 2D survival game specifically designed to test long-horizon reasoning and sequential decision-making, with tasks involving resource gathering, crafting, and combat. It represents a significantly more complex setting than \alfworld{} and \webshop{}. 

We evaluated our method on this benchmark using \llamaThreeOneEightIt{}, following \balrog{}'s official evaluation protocol. We compare \name{} with several representative baselines from \balrog{}'s GitHub repository. The preliminary results are summarized in \cref{tab:crafter}. Our method achieves higher performance than other available baselines under identical evaluation settings. These preliminary results provide further evidence of the robustness, effectiveness, and adaptability of our proposed actor-critic approach, particularly in significantly more challenging and complex decision-making environments. It is worth noting that CoT performs worse than the baseline Naive on the \crafter{} benchmark. 
We hypothesize that it is due to the model's inconsistency within their chain-of-thoughts. Two contiguous chains of thoughts might lead model to take actions which push towards different goals, which is not ideal. The authors also note that this is a problem especially with smaller, weaker models.

\begin{table}[ht]
\centering
\caption{Performance comparison of \name{} (Ours) and other baselines in \crafter{} (from \balrog{})}\label{tab:crafter}
\begin{tabular}{@{}lc@{}}
\toprule
                                        & \multicolumn{1}{l}{\llamaThreeOneEightIt{}}                        \\ \midrule
\name{} (Ours)                             & \textbf{25.91\% ± 1.93\%}                                       \\
Naive (direct action generation)        & 20.45\% ± 4.26\%                                                \\
Robust Naive (formatted actions)        & 4.55\% ± 1.57\%                                                 \\
CoT (\react{}, reason then act)            & 18.64\% ± 3.24\%                                                \\
Robust CoT (reason + formmated actions) & 15.46\% ± 3.59\%                                                \\
Few-Shot (in-context examples)          & 12.73\% ± 1.25\% \\ \bottomrule
\end{tabular}
\end{table}

\subsection{Preliminary experiments with reasoning LLMs}
We have conducted preliminary experiments with reasoning LLMs like DeepSeek-R1-Distill-Qwen-7B \cite{deepseekai2025deepseekr1}. However, we observed that they often tend to overthink rather than output direct environmental actions in both our method and the baseline \react{}. For instance, even when we explicitly prompt the reasoning LLMs to output actions (\eg{}, ``Please make sure your response is in the following format:\textbackslash n> \{The action you choose\}''), the models still generated detailed explanations but avoided selecting the next action. A typical response might be: ``I need to find a key to open the safe or locate the pencil in the drawers. Since I can't (…), I'm unable (…), I must (…)''. This issue has also been noted in prior work \cite{cuadron2025danger}. We believe that using reasoning LLMs for decision-making tasks requires deeper exploration to balance internal reasoning with effective environmental interaction.
\section{Method details}\label{appx:method}

\subsection{Deriving the solution of the KL-constrained maximization objective}\label{appx:derive_solution}
In this subsection, we will derive \cref{equ:solution}. We optimize the following objective:
\begin{equation*}
    \max_\pi \mathbb{E}_{a_t^i\sim \pi(a_t^i|g,h_t)}\left[\gVL(g,h_t,a_t^i,u_t^i)\right]-\frac{1}{\alpha} \mathbb{D}_{KL}\left[\pi(a_t^i|g,h_t)\| \piL(a_t^i|g,h_t)\right].
\end{equation*}
We now have:
\begin{align*}
    &\max_\pi \mathbb{E}_{a_t^i\sim \pi(a_t^i|g,h_t)}\left[\gVL(g,h_t,a_t^i,u_t^i)\right]-\frac{1}{\alpha} \mathbb{D}_{KL}\left[\pi(a_t^i|g,h_t)\| \piL(a_t^i|g,h_t)\right] \\
    =& \max_\pi \mathbb{E}_{a_t^i\sim \pi(a_t^i|g,h_t)}\left[\gVL(g,h_t,a_t^i,u_t^i) - \frac{1}{\alpha}\log\frac{\pi(a_t^i|g,h_t)}{\piL(a_t^i|g,h_t)} \right]\\
    =& \min_\pi \mathbb{E}_{a_t^i\sim \pi(a_t^i|g,h_t)}\left[\log\frac{\pi(a_t^i|g,h_t)}{\piL(a_t^i|g,h_t)} - \alpha \gVL(g,h_t,a_t^i,u_t^i)\right]\\
    =& \min_\pi \mathbb{E}_{a_t^i\sim \pi(a_t^i|g,h_t)}\left[\log\frac{\pi(a_t^i|g,h_t)}{\frac{1}{Z(g,h_t)}\piL(a_t^i|g,h_t)\exp{(\alpha \gVL(g,h_t,a_t^i,u_t^i))}} - \log Z(g,h_t) \right]\\
\end{align*}
where we have the partition function:
\begin{equation*}
    Z(g,h_t)=\sum_{a_t^i} \piL(a_t^i|g,h_t)\exp{(\alpha \gVL(g,h_t,a_t^i,u_t^i))}.
\end{equation*}
Since the partition function is a function of only $g,h_t$ and the original policy $\piL$, but does not depend on the optimized policy $\pi$, we define
\begin{equation*}
    \pi^*(a_t^i|g,h_t)=\frac{1}{Z(g,h_t)}\piL(a_t^i|g,h_t)\exp{(\alpha \gVL(g,h_t,a_t^i,u_t^i))}.
\end{equation*}
This definition of policy if a valid probability distribution as $\pi^*(a_t^i|g,h_t)$ for all $a_t^i$ and $\sum_{a_t^i}\pi^*(a_t^i|g,h_t)=1$. As $Z(g,h_t)$ is not a function of $a_t^i$, we can then re-organize the objective as:
\begin{align*}
    &\min_\pi \mathbb{E}_{a_t^i\sim \pi(a_t^i|g,h_t)}\left[\log\frac{\pi(a_t^i|g,h_t)}{\frac{1}{Z(g,h_t)}\piL(a_t^i|g,h_t)\exp{(\alpha \gVL(g,h_t,a_t^i,u_t^i))}} - \log Z(g,h_t) \right]\\
    =& \min_\pi \mathbb{E}_{a_t^i\sim \pi(a_t^i|g,h_t)}\left[\log\frac{\pi(a_t^i|g,h_t)}{\pi^*(a_t^i|g,h_t)} - \log Z(g,h_t) \right] \\
    =& \min_\pi \mathbb{D}_{KL}\left[\pi(a_t^i|g,h_t)\|\pi^*(a_t^i|g,h_t)\right] - \log Z(g,h_t).
\end{align*}

Then since $Z(g,h_t)$ does not depend on $\pi$, we can only care about the KL-divergence, which is minimized at 0 if and only if the two distributions are identical. Therefore, the optimal solution is
\begin{equation}
    \pi(a_t^i|g,h_t)=\pi^*(a_t^i|g,h_t)=\frac{1}{Z(g,h_t)}\piL(a_t^i|g,h_t)\exp{(\alpha \gVL(g,h_t,a_t^i,u_t^i))},
\end{equation}
which completes the derivation.


\subsection{Critic improvement of \name{}}\label{appx:critic_improve}

The reflections critic $\gVL$, and forward model $\fL$ we used can be easily implemented by prompting LLMs via providing instructions or few-shot examples from similar tasks like prior work \citep{yao2022react, liu2023rafa}. However, empirically, we found that they can be further improved via fine-tuning LLMs with simple next-token prediction loss on several samples collected from training tasks. In this work, we consider fine-tuning for \alfworld{} and \babyai{} to show the impact of fine-tuning. We use 18 trajectories for each benchmark for fine-tuning, including two successful trajectories and one failed trajectory for each task-type. Though 18 trajectories are significantly fewer than what is required for conventional reinforcement learning algorithms, they are generally enough for our method. Each trajectory has the following format: $(g,o_{0},a_{1},o_{1},c_{1},\cdots,a_{H},o_{H},c_{H})$, where $H$ is the episode length and $c_i$ is a reflection of action $a_t$. Each $c_t$ includes an explanation about the action $a_t$ (\eg{}, ``I have found object-X. This step is '' or ``I should take object-X instead of object-Y first. This step is '') and a special token that indicates positive/negative judgment (\eg{}, ``GOOD'' or ``BAD''). 

Practically, we just fine-tune the LLM once and use it to generate all the reflections, construct critic $\gVL$, and forward model $\fL$, thanks to the fine-tuning with the above data format. Specifically, when minimizing the loss of predicting future trajectories, the forward model $\fL$ is improved. When minimizing the loss of generating reflections, the reflections, critic $\gVL$ are both improved. The latter is because reflections also contain special tokens that indicate positive/negative judgments, whose generated probabilities are used to calculate $\gVL$ in \cref{equ:def_q_future}.  We analyze this fine-tuning process in \cref{appx:finetuning-analysis}. Some examples of the labeled trajectories in \alfworld{} and \babyai{} are shown in \cref{tab:alfworld_ours_example} and \cref{tab:babyai_ours_example} respectively.

\section{Experiment details}\label{appx:exp}

\subsection{Benchmark details}

\subsubsection{\alfworld{}: benchmark with high-level actions}
We choose \alfworld{} \citep{shridhar2021alfworld}, a text-based household environment, to demonstrate the effectiveness of \name{} on high-level planning. \alfworld{} is a synthetic text-based game aligned with \alfred{} \citep{shridhar2020alfred} benchmark. There are 6 types of tasks in this environment, which require the agent to achieve a high-level goal through a sequence of high-level actions, \eg{} ``go to place-X'', ``take object-Y from place-X'', \etc{} The details about the 6 task types in \alfworld{} are shown in \cref{tab:alfworld_task_type}.


\begin{table}[htbp]
\scriptsize
    \centering
    \caption{All the task types and the corresponding goals for \alfworld{}}
    \begin{tabular}{l p{0.75\linewidth}}
        \toprule
        Type & Description \\ \midrule
        Pick \& Place & The agent needs to put a target object to a target place, \eg{} put some spraybottle on toilets, find some apple and put it in sidetable, \etc{} \\ 
        Clean \& Place & The agent needs to find a target object, clean it and put it to a target place, \eg{} clean some apple and put it in sidetable, put a clean lettuce in diningtable, \etc{} \\ 
        Heat \& Place & The agent needs to find a target object, heat it and put it to a target place, \eg{} heat some egg and put it in diningtable, put a hot apple in fridge, \etc{} \\ 
        Cool \& Place & The agent needs to find a target object, cool it and put it to a target place, \eg{} cool some pan and put it in stoveburner, put a cool mug in shelf, \etc{} \\ 
        Examine \& Place & The agent needs to find a target object, and examine it with desklamp, \eg{} look at bowl under the desklamp, examine the pen with the desklamp, \etc{} \\ 
        Pick Two \& Place & The agent needs to put two target objects to a target place, \eg{} put two saltshaker in drawer, find two pen and put them in dresser, \etc{} \\ \bottomrule
    \end{tabular}
    \label{tab:alfworld_task_type}
\end{table}

A challenge built into \alfworld{} is that the agent needs to explore the environment to find a target object. The commonsense knowledge in LLMs about the likely locations for common household items makes this environment suitable for LLMs to solve. The reward is 1 only when the agent reaches the goal. Following \react{}, we evaluate 134 unseen evaluation games in a task-specific setup.

\subsubsection{\babyai{}: benchmark with low-level actions}

For decision-making tasks with low-level planning, we adopt \babyai{} \citep{carta2023grounding} as our test-bed. \babyai{} is a text-only version environment extended from the BabyAI platform \citep{chevalierboisvert2019babyai}. \babyai{} is a Grid World environment, in which the agent and objects are placed in a room of $8\times8$ tiles. The agent has 6 primitive actions: turn left, turn right, go forward, pick up, drop, toggle, to solve a task described in natural language (\eg{} Pick up the red box). The agent has access to a $7\times7$ partial view, which means it can only observe the objects belonging to the $7\times7$ grid in front of it. In addition to objects relevant to completing a given task, there are also other distractors in the room. All the task types in \babyai{} are shown in \cref{tab:babyai_task_type}.


\begin{table}[htbp]
\scriptsize
    \centering
    \caption{All the task types and the corresponding goals for \babyai{}}
    \begin{tabular}{l p{0.75\linewidth}}
        \toprule
        Type & Description \\ \midrule
        go to & The agent needs to find target object and go to it, \eg{} go to the green key, go to the red ball, \etc{} \\ 
        pick up & The agent needs to find target object, go to it and pick up it, \eg{} pick up the blue key, pick up the purple ball, \etc{} \\ 
        go to after pick up & The agent needs to find and pick up one object, then go to another object, \eg{} go to the blue key after you pick up the green key \etc{} \\ 
        pick up then go to & The agent needs to find and pick up one object, then go to another object, \eg{} pick up the green box, then go to the purple box \etc{} \\ 
        put next to & The agent needs to find and pick up one object, then go to another object and put the first object next to it, \eg{} put the grey key next to the yellow ball \etc{} \\ 
        open door & The agent needs to know which key to pick up, then find and pick up it to open the door, \eg{} open the door, open the blue door, \etc{} \\ \bottomrule
    \end{tabular}
    \label{tab:babyai_task_type}
\end{table}

Unlike \alfworld{}, an agent interacting with \babyai{} needs to find out the suitable low-level action to execute at each step. We evaluate on the test environment in \babyai{}. The objects in a room are randomly chosen, and their position, as well as the agent's position, are also random. Considering the time and computational resource constraints, we evaluate on 50 tasks for each task type, yielding 300 tasks total.

\subsubsection{\webshop{}: benchmark with potentially infinite action space}

In some scenarios, the space of possible actions can be infinite. We adopt a simplified environment, \webshop{} \citep{yao2022webshop}, with potentially infinite action space to demonstrate the effectiveness of our method. \webshop{} requires an agent to purchase a desired product based on textual instructions through the interaction with a simulated web environment. The available actions include search something using textual descriptions and choose button to click like "[item ID]", "back to search", "buy now", \etc{} Due to the nature of natural language, one can actually search anything, leading to an unbounded set of potential actions. We evaluate 100 tasks in \webshop{} for our method and all the baselines.

One thing that should be mentioned about this benchmark is that the final reward is a value between 0 and 1, depending on the degree to which the final purchased product meets the requirements. Thus in this work, we report the success rate and final reward simultaneously, where success rate indicates the proportion of products that the agent finally buys that fully meet the requirements, and final reward indicates the average final reward on all evaluated tasks.

\subsection{Implementation details of baselines}\label{appx:exp_baseline}

\subsubsection{\react{}}

\react{} \citep{yao2022react} uses LLM as policy to generate action directly. They combine reasoning and acting during the interaction with the environment to improve action selection.

\cref{tab:alfworld_act_example} shows an example of a trajectory interleaving between actions and observations in \alfworld{}. To leverage the reasoning ability of LLMs, \react{} adds an extra thinking action to the original action set in \alfworld{}, allowing LLM agent to engage in reasoning about task decomposition and goal achievement during task execution. An annotated trajectory for \alfworld{} given by \react{} is shown in \cref{tab:alfworld_react_example}. Text labeled in light green is the annotated thought to help LLM complete a task better.

\begin{table}[htbp]
\scriptsize
    \caption{An example of successful trajectory for \alfworld{}}
    \centering
    \ttfamily
    \begin{tabular}{p{0.9\linewidth}}
        \toprule
You are in the middle of a room. Looking quickly around you, you see a cabinet 4, a cabinet 3, a cabinet 2, a cabinet 1, a countertop 1, a garbagecan 1, a handtowelholder 2, a handtowelholder 1, a sinkbasin 2, a sinkbasin 1, a toilet 1, a toiletpaperhanger 1, and a towelholder 1.

Your task is to: put some spraybottle on toilet.

> go to cabinet 1

On the cabinet 1, you see a cloth 1, a soapbar 1, a soapbottle 1.

> go to cabinet 2

The cabinet 2 is closed.

> open cabinet 2

You open the cabinet 2. The cabinet 2 is open. In it, you see a candle 1, and a spraybottle 2.

> take spraybottle 2 from cabinet 2

You pick up the spraybottle 2 from the cabinet 2.

> go to toilet 1

On the toilet 1, you see a soapbottle 2.

> put spraybottle 2 in/on toilet 1

You put the spraybottle 2 in/on the toilet 1.\\ \bottomrule
    \end{tabular}
    \label{tab:alfworld_act_example}
\end{table}

\begin{table}[htbp]
\scriptsize
    \caption{An example of annotated successful trajectory for \alfworld{} in \react{}}
    \centering
    \ttfamily
    \begin{tabular}{p{0.9\linewidth}}
        \toprule
You are in the middle of a room. Looking quickly around you, you see a cabinet 4, a cabinet 3, a cabinet 2, a cabinet 1, a countertop 1, a garbagecan 1, a handtowelholder 2, a handtowelholder 1, a sinkbasin 2, a sinkbasin 1, a toilet 1, a toiletpaperhanger 1, and a towelholder 1.

Your task is to: put some spraybottle on toilet.

\textcolor{thought}{> think: To solve the task, I need to find and take a sparybottle, then put it on toilet.}

OK.

\textcolor{thought}{> think: First I need to find a spraybottle. A spraybottle is more likely to appear in cabinet (1-4), countertop (1), toilet (1), sinkbasin (1-2), garbagecan (1). I can check one by one, starting with cabinet 1.}

OK.

> go to cabinet 1

On the cabinet 1, you see a cloth 1, a soapbar 1, a soapbottle 1.

> go to cabinet 2

The cabinet 2 is closed.

> open cabinet 2

You open the cabinet 2. The cabinet 2 is open. In it, you see a candle 1, and a spraybottle 2.

\textcolor{thought}{> think: Now I find a spraybottle (2). Next, I need to take it.}

> take spraybottle 2 from cabinet 2

You pick up the spraybottle 2 from the cabinet 2.

\textcolor{thought}{> think: Now I take a spraybottle (2). Next, I need to put it in/on toilet 1.}

OK.

> go to toilet 1

On the toilet 1, you see a soapbottle 2.

> put spraybottle 2 in/on toilet 1

You put the spraybottle 2 in/on the toilet 1.\\ \bottomrule
    \end{tabular}
    \label{tab:alfworld_react_example}
\end{table}

During the interaction with the environment, two successful trajectories like in \cref{tab:alfworld_react_example} are provided to LLMs to leverage the in-context learning ability of LLMs. The LLMs will generate an action to be executed in the environment or conduct some reasoning about how to achieve the final goal. The chosen action and new observation are appended to the prompt for LLMs to form a sequential decision-making process.

\begin{table}[htbp]
\scriptsize
    \caption{An example of successful trajectory for \babyai{}}
    \centering
    \ttfamily
    \begin{tabular}{p{0.9\linewidth}}
        \toprule
Goal of the agent: go to the green key \\
Observation:You see a wall 2 steps right, You see a wall 3 steps forward, You see a grey box 3 steps left, You see a grey key 3 steps left and 1 step forward, You see a blue box 3 steps left and 2 steps forward \\
Action:turn left \\
Observation:You see a wall 3 steps right, You see a blue key 3 steps left and 2 steps forward, You see a green key 2 steps left and 1 step forward, You see a green ball 1 step left and 3 steps forward, You see a grey box 3 steps forward, You see a grey key 1 step right and 3 steps forward, You see a blue box 2 steps right and 3 steps forward \\
Action:go forward \\
Observation:You see a wall 3 steps right, You see a blue key 3 steps left and 1 step forward, You see a green key 2 steps left, You see a green ball 1 step left and 2 steps forward, You see a grey box 2 steps forward, You see a grey key 1 step right and 2 steps forward, You see a blue box 2 steps right and 2 steps forward \\
Action:turn left \\
Observation:You see a wall 3 steps left, You see a green box 2 steps left and 3 steps forward, You see a green box 2 steps left and 1 step forward, You see a green key 2 steps forward, You see a blue key 1 step right and 3 steps forward, You see a green ball 2 steps right and 1 step forward, You see a grey box 2 steps right \\
Action:go forward \\
Observation:You see a green box 2 steps left and 2 steps forward, You see a green box 2 steps left, You see a green key 1 step forward, You see a blue key 1 step right and 2 steps forward, You see a green ball 2 steps right\\ \bottomrule
    \end{tabular}
    \label{tab:babyai_act_example}
\end{table}

\begin{table}[htbp]
\scriptsize
    \caption{An example of annotated successful trajectory for \babyai{} in \react{}}
    \centering
    \ttfamily
    \begin{tabular}{p{0.9\linewidth}}
        \toprule
Goal of the agent: go to the green key \\
Observation:You see a wall 2 steps right, You see a wall 3 steps forward, You see a grey box 3 steps left, You see a grey key 3 steps left and 1 step forward, You see a blue box 3 steps left and 2 steps forward \\
\textcolor{thought}{Action:think: To complete the task, I need to find a green key and go to it.} \\
Observation:OK \\
\textcolor{thought}{Action:think: I have not seen a green key. Firstly, I need to find a green key.} \\
Observation:OK \\
Action:turn left \\
Observation:You see a wall 3 steps right, You see a blue key 3 steps left and 2 steps forward, You see a green key 2 steps left and 1 step forward, You see a green ball 1 step left and 3 steps forward, You see a grey box 3 steps forward, You see a grey key 1 step right and 3 steps forward, You see a blue box 2 steps right and 3 steps forward \\
\textcolor{thought}{Action:think: I have seen a green key 2 steps left and 1 step forward. Next, I need to go to the green key.} \\
Observation:OK \\
Action:go forward \\
Observation:You see a wall 3 steps right, You see a blue key 3 steps left and 1 step forward, You see a green key 2 steps left, You see a green ball 1 step left and 2 steps forward, You see a grey box 2 steps forward, You see a grey key 1 step right and 2 steps forward, You see a blue box 2 steps right and 2 steps forward \\
Action:turn left \\
Observation:You see a wall 3 steps left, You see a green box 2 steps left and 3 steps forward, You see a green box 2 steps left and 1 step forward, You see a green key 2 steps forward, You see a blue key 1 step right and 3 steps forward, You see a green ball 2 steps right and 1 step forward, You see a grey box 2 steps right \\
Action:go forward \\
Observation:You see a green box 2 steps left and 2 steps forward, You see a green box 2 steps left, You see a green key 1 step forward, You see a blue key 1 step right and 2 steps forward, You see a green ball 2 steps right\\ \bottomrule
    \end{tabular}
    \label{tab:babyai_react_example}
\end{table}

Similar to \alfworld{}, we annotate some successful trajectories in \babyai{} to teach a \react{} agent how to complete a task in \babyai{}. Examples of successful trajectories without thoughts and with thoughts are shown in \cref{tab:babyai_act_example} and \cref{tab:babyai_react_example} respectively.

\begin{table}[htbp]
\scriptsize
    \caption{An example of successful trajectory for \webshop{}}
    \centering
    \ttfamily
    \begin{tabular}{p{0.9\linewidth}}
        \toprule
Webshop \\
Instruction:  \\
i would like a 3 ounce bottle of bright citrus deodorant for sensitive skin, and price lower than 50.00 dollars \\
\text{[Search]}  \\
\\
Action: search[3 ounce bright citrus deodorant sensitive skin] \\
Observation: \\
\text{[Back to Search]} \\
Page 1 (Total results: 50) \\
\text{[Next >]} \\
\text{[B078GWRC1J]} \\
Bright Citrus Deodorant by Earth Mama | Natural and Safe for Sensitive Skin, Pregnancy and Breastfeeding, Contains Organic Calendula 3-Ounce \\
\$10.99 \\
\text{[B078GTKVXY]} \\ 
Ginger Fresh Deodorant by Earth Mama | Natural and Safe for Sensitive Skin, Pregnancy and Breastfeeding, Contains Organic Calendula 3-Ounce \\
\$10.99 \\
\text{[B08KBVJ4XN]} \\
Barrel and Oak - Aluminum-Free Deodorant, Deodorant for Men, Essential Oil-Based Scent, 24-Hour Odor Protection, Cedar \& Patchouli Blend, Gentle on Sensitive Skin (Mountain Sage, 2.7 oz, 2-Pack) \\
\$15.95  \\
\\
Action: click[B078GWRC1J] \\
Observation: \\
\text{[Back to Search]} \\
\text{[< Prev]} \\
scent [assorted scents][bright citrus][calming lavender][ginger fresh][simply non-scents] \\
size [travel set (4-pack)][3 ounce (pack of 1)][3-ounce (2-pack)] \\
Bright Citrus Deodorant by Earth Mama | Natural and Safe for Sensitive Skin, Pregnancy and Breastfeeding, Contains Organic Calendula 3-Ounce \\
Price: \$10.99 \\
Rating: N.A. \\
\text{[Description]} \\
\text{[Features]} \\
\text{[Reviews]} \\
\text{[Buy Now]}  \\
\\
Action: click[bright citrus] \\
Observation: You have clicked bright citrus. \\
\\
Action: click[3 ounce (pack of 1)] \\
Observation: You have clicked 3 ounce (pack of 1). \\
\\
Action: click[Buy Now] \\
Observation: Your score (min 0.0, max 1.0): 1.0 \\ \bottomrule
    \end{tabular}
    \label{tab:webshop_act_example}
\end{table}

\begin{table}[htbp]
\scriptsize
    \caption{An example of successful trajectory for \webshop{} in \react{}}
    \centering
    \ttfamily
    \begin{tabular}{p{0.9\linewidth}}
        \toprule
Webshop \\
Instruction:  \\
i would like a 3 ounce bottle of bright citrus deodorant for sensitive skin, and price lower than 50.00 dollars \\
\text{[Search]}  \\
\\
Action: search[3 ounce bright citrus deodorant sensitive skin] \\
Observation: \\
\text{[Back to Search]} \\
Page 1 (Total results: 50) \\
\text{[Next >]} \\
\text{[B078GWRC1J]} \\
Bright Citrus Deodorant by Earth Mama | Natural and Safe for Sensitive Skin, Pregnancy and Breastfeeding, Contains Organic Calendula 3-Ounce \\
\$10.99 \\
\text{[B078GTKVXY]} \\ 
Ginger Fresh Deodorant by Earth Mama | Natural and Safe for Sensitive Skin, Pregnancy and Breastfeeding, Contains Organic Calendula 3-Ounce \\
\$10.99 \\
\text{[B08KBVJ4XN]} \\
Barrel and Oak - Aluminum-Free Deodorant, Deodorant for Men, Essential Oil-Based Scent, 24-Hour Odor Protection, Cedar \& Patchouli Blend, Gentle on Sensitive Skin (Mountain Sage, 2.7 oz, 2-Pack) \\
\$15.95  \\
\\
\textcolor{thought}{Action: think[[B078GWRC1J] and [B078GTKVXY] are bright citrus deodorant less then 50 dollars. I can check [B078GWRC1J] first.]} \\
Observation: OK. \\
\\
Action: click[B078GWRC1J] \\
Observation: \\
\text{[Back to Search]} \\
\text{[< Prev]} \\
scent [assorted scents][bright citrus][calming lavender][ginger fresh][simply non-scents] \\
size [travel set (4-pack)][3 ounce (pack of 1)][3-ounce (2-pack)] \\
Bright Citrus Deodorant by Earth Mama | Natural and Safe for Sensitive Skin, Pregnancy and Breastfeeding, Contains Organic Calendula 3-Ounce \\
Price: \$10.99 \\
Rating: N.A. \\
\text{[Description]} \\
\text{[Features]} \\
\text{[Reviews]} \\
\text{[Buy Now]}  \\
\\
\textcolor{thought}{Action: think[For 3 ounce bottle of bright citrus deodorant for sensitive skin, the item has options [bright citrus] and [3 ounce (pack of 1)] and seems good to buy.]} \\
Observation: OK. \\
\\
Action: click[bright citrus] \\
Observation: You have clicked bright citrus. \\
\\
Action: click[3 ounce (pack of 1)] \\
Observation: You have clicked 3 ounce (pack of 1). \\
\\
Action: click[Buy Now] \\
Observation: Your score (min 0.0, max 1.0): 1.0 \\ \bottomrule
    \end{tabular}
    \label{tab:webshop_react_example}
\end{table}

As to \webshop{}, we adopt the implementation of \react{} and use one example for in-context learning. The successful example without thoughts and with thoughts are shown in \cref{tab:webshop_act_example} and \cref{tab:webshop_react_example} respectively.

\subsubsection{\rap{}}

\rap \citep{hao2023rap} utilizes LLMs as policy and world model, and adopts tree-search planning methods to evaluate each possible action candidate. Trajectories interleaving between actions and observations like in \cref{tab:alfworld_act_example}, \cref{tab:babyai_act_example} and \cref{tab:webshop_act_example} are provided to LLMs to learn how to choose action as an policy and the dynamics of the environment as a world model.

The assessment of each step is performed by a reward function, which can be the log probability of the action or self-evaluation given by LLMs, or based on some task-specific heuristics. In our implementation, we adopt the log probability of actions given by LLMs as the reward. For simplicity of implementation, we adopted a greedy approach to expand the tree, generating only one action at a time. More specifically, at each step, LLMs will sample some action candidates. For each action candidate, LLMs will generate a rollout trajectory until a maximum step or terminal state. The summation of log probabilities of all the actions on the rollout accessed by LLMs are used as Q value for each action candidate. The candidate with the highest Q value is chosen to be executed in the environment.

\subsubsection{\icpi{}}

\begin{table}[htbp]
\scriptsize
    \caption{An example provided to critic in \icpi{} for \alfworld{}}
    \centering
    \ttfamily
    \begin{tabular}{p{0.9\linewidth}}
        \toprule
You are in the middle of a room. Looking quickly around you, you see a cabinet 4, a cabinet 3, a cabinet 2, a cabinet 1, a countertop 1, a garbagecan 1, a handtowelholder 2, a handtowelholder 1, a sinkbasin 2, a sinkbasin 1, a toilet 1, a toiletpaperhanger 1, and a towelholder 1.

Your task is to: put some spraybottle on toilet.

> go to cabinet 1

On the cabinet 1, you see a cloth 1, a soapbar 1, a soapbottle 1.

\textcolor{critic}{Reward:0}

> go to cabinet 2

The cabinet 2 is closed.

\textcolor{critic}{Reward:0}

> open cabinet 2

You open the cabinet 2. The cabinet 2 is open. In it, you see a candle 1, and a spraybottle 2.

\textcolor{critic}{Reward:0}

> take spraybottle 2 from cabinet 2

You pick up the spraybottle 2 from the cabinet 2.

\textcolor{critic}{Reward:0}

> go to toilet 1

On the toilet 1, you see a soapbottle 2.

\textcolor{critic}{Reward:0}

> put spraybottle 2 in/on toilet 1

You put the spraybottle 2 in/on the toilet 1.

\textcolor{critic}{Reward:1}\\ \bottomrule
    \end{tabular}
    \label{tab:alfworld_icpi_example}
\end{table}

\begin{table}[htbp]
\scriptsize
    \caption{An example provided to critic in \icpi{} for \babyai{}}
    \centering
    \ttfamily
    \begin{tabular}{p{0.9\linewidth}}
        \toprule
Goal of the agent: go to the green key

Observation:You see a wall 2 steps right, You see a wall 3 steps forward, You see a grey box 3 steps left, You see a grey key 3 steps left and 1 step forward, You see a blue box 3 steps left and 2 steps forward

Action:turn left

Observation:You see a wall 3 steps right, You see a blue key 3 steps left and 2 steps forward, You see a green key 2 steps left and 1 step forward, You see a green ball 1 step left and 3 steps forward, You see a grey box 3 steps forward, You see a grey key 1 step right and 3 steps forward, You see a blue box 2 steps right and 3 steps forward

\textcolor{critic}{Reward:0}

Action:go forward

Observation:You see a wall 3 steps right, You see a blue key 3 steps left and 1 step forward, You see a green key 2 steps left, You see a green ball 1 step left and 2 steps forward, You see a grey box 2 steps forward, You see a grey key 1 step right and 2 steps forward, You see a blue box 2 steps right and 2 steps forward

\textcolor{critic}{Reward:0}

Action:turn left

Observation:You see a wall 3 steps left, You see a green box 2 steps left and 3 steps forward, You see a green box 2 steps left and 1 step forward, You see a green key 2 steps forward, You see a blue key 1 step right and 3 steps forward, You see a green ball 2 steps right and 1 step forward, You see a grey box 2 steps right

\textcolor{critic}{Reward:0}

Action:go forward

Observation:You see a green box 2 steps left and 2 steps forward, You see a green box 2 steps left, You see a green key 1 step forward, You see a blue key 1 step right and 2 steps forward, You see a green ball 2 steps right

\textcolor{critic}{Reward:1}\\ \bottomrule
    \end{tabular}
    \label{tab:babyai_icpi_example}
\end{table}

\icpi{} \citep{brooks2024large} proposes to implement policy iteration using LLMs through in-context learning. At each step, the policy in \icpi{} will sample some action candidates and the critic will compute the Q values for each action candidates. The action candidates with the highest Q values is chosen to be executed.

The policy is implemented using LLMs, and successful trajectories like in \cref{tab:alfworld_act_example} and \cref{tab:babyai_act_example} are provided to it.

As to the critic, \icpi{} prompts LLMs to give the numerical reward for each step directly. Given the current history and an action candidate, the critic in \icpi{} will rollout a trajectory starting from the action candidate. Apart from predicting the observations, the critic will give the numerical reward for each step on the rollout trajectory. The (discounted) return on the rollout is treated as the Q value for the action candidate. For both \alfworld{} and \babyai{}, we define the reward as 1 when the agent reaches the goal. All other steps will have a reward 0. The examples provided to the critic are like in \cref{tab:alfworld_icpi_example} for \alfworld{} and \cref{tab:babyai_icpi_example} for \babyai{}.

\subsubsection{\rafa{}}

\begin{table}[htbp]
\scriptsize
    \caption{An example provided to critic in \rafa{} for \alfworld{}}
    \centering
    \ttfamily
    \begin{tabular}{p{0.9\linewidth}}
        \toprule
You are in the middle of a room. Looking quickly around you, you see a cabinet 4, a cabinet 3, a cabinet 2, a cabinet 1, a countertop 1, a garbagecan 1, a handtowelholder 2, a handtowelholder 1, a sinkbasin 2, a sinkbasin 1, a toilet 1, a toiletpaperhanger 1, and a towelholder 1.

Your task is to: put some spraybottle on toilet.

\textcolor{critic}{> critic: My task requires two sub-goals in order: take a spraybottle and put the spraybottle on the toilet. My current state satisfies zero of the two sub-goals. The value is 0/2=0.}

> OK.

On the cabinet 1, you see a cloth 1, a soapbar 1, a soapbottle 1.

> OK.

The cabinet 2 is closed.

> OK.

You open the cabinet 2. The cabinet 2 is open. In it, you see a candle 1, and a spraybottle 2.

> OK.

You pick up the spraybottle 2 from the cabinet 2.

\textcolor{critic}{> critic: Now I take a spraybottle. My current state satisfies the first of the two sub-goals: take a spraybottle. The value is 1/2=0.5.}

> OK.

On the toilet 1, you see a soapbottle 2.

> OK.

You put the spraybottle 2 in/on the toilet 1.

\textcolor{critic}{> critic: Now I put the spraybottle on the toilet. My current state satisfies all the two sub-goals. The value is 2/2=1.}\\ \bottomrule
    \end{tabular}
    \label{tab:alfworld_rafa_example}
\end{table}

\begin{table}[htbp]
\scriptsize
    \caption{An example provided to critic in \rafa{} for \babyai{}}
    \centering
    \ttfamily
    \begin{tabular}{p{0.9\linewidth}}
        \toprule
Goal of the agent: go to the green key

You see a wall 2 steps right, You see a wall 3 steps forward, You see a grey box 3 steps left, You see a grey key 3 steps left and 1 step forward, You see a blue box 3 steps left and 2 steps forward

\textcolor{critic}{>critic: My task requires two sub-goals in order: find the green key, and go to the green key. My current state satisfies zero of the two sub-goals. The value is 0/2=0.}

>OK.

You see a wall 3 steps right, You see a blue key 3 steps left and 2 steps forward, You see a green key 2 steps left and 1 step forward, You see a green ball 1 step left and 3 steps forward, You see a grey box 3 steps forward, You see a grey key 1 step right and 3 steps forward, You see a blue box 2 steps right and 3 steps forward

\textcolor{critic}{>critic: Now I find the green key. My current state satisfies the first of the two sub-goals: find the green key. The value is 1/2=0.5.}

>OK.

You see a wall 3 steps right, You see a blue key 3 steps left and 1 step forward, You see a green key 2 steps left, You see a green ball 1 step left and 2 steps forward, You see a grey box 2 steps forward, You see a grey key 1 step right and 2 steps forward, You see a blue box 2 steps right and 2 steps forward

>OK.

You see a wall 3 steps left, You see a green box 2 steps left and 3 steps forward, You see a green box 2 steps left and 1 step forward, You see a green key 2 steps forward, You see a blue key 1 step right and 3 steps forward, You see a green ball 2 steps right and 1 step forward, You see a grey box 2 steps right

>OK.

You see a green box 2 steps left and 2 steps forward, You see a green box 2 steps left, You see a green key 1 step forward, You see a blue key 1 step right and 2 steps forward, You see a green ball 2 steps right

\textcolor{critic}{>critic: Now I go to the green key. My current state satisfies all the two sub-goals. The value is 2/2=1.}\\ \bottomrule
    \end{tabular}
    \label{tab:babyai_rafa_example}
\end{table}

The framework of \rafa{} \citep{liu2023rafa} is also like \rap{} or \icpi{}. The main difference is how the action evaluation is conducted.

\rafa{} implements tree-search using LLM to evaluate each action candidate. Different from \icpi{}, \rafa{} uses the task completion progress as the value for each step. They have the LLMs decompose a goal into sub-goals, and use the completion status of the sub-goals after each step as the value for the step. \rafa{} evaluates the completion status of sub-goals based on the predicted observations. Examples provided in \rafa{} are like in \cref{tab:alfworld_rafa_example} for \alfworld{} and \cref{tab:babyai_rafa_example} for \babyai{}.

\subsubsection{\lats{}}

\lats{} \citep{zhou2024lats} combines the reasoning, acting, and planning capabilities of LLMs with MCTS \citep{kocsis2006bandit} and external feedback mechanisms to enhance decision-making, achieving competitive results in web navigation. We adopt the official implementation to evaluate its performance on \webshop{}.

\lats{} cannot be directly applied to the other benchmarks we used for two main reasons. (1) \lats{} requires the ability to revert the agent to earlier states in the environment, which \alfworld{} and \babyai{} do not support. \lats{} relies on model-free MCTS, using environment simulator as a world model, reverting simulators to earlier states during tree search. This limitation is also noted in their original paper (Page 9). (2) While it might be possible to modify these environments to make them reversible, it would create an unfair comparison. Our method and other baselines do not rely on simulators during reasoning in \alfworld{} and \babyai{}, whereas, \lats{} would gain an advantage from this modification.

Nevertheless, we still attempted to adapt \lats{} for \alfworld{} by using LLMs as world models, similar to our method, for a fair comparison. The results, presented in the \cref{tab:lats}, show that \lats{} fails in almost all tasks. This is because its tree search severely depends on the environment simulator for precise state transitions. With only LLM-based world models, the state transitions often deviate from the actual environments, due to LLMs' inherent hallucinations and partial observability of \alfworld{}.

\begin{table}[ht]
\centering
\caption{Performance comparison of \name{} (Ours) and \lats{} in \alfworld{}.}\label{tab:lats}
\begin{tabular}{@{}lcccc@{}}
\toprule
{ }      & \multicolumn{1}{l}{\codellamaSeven{}} & \multicolumn{1}{l}{\gemmaSeven{}} & \multicolumn{1}{l}{\llamaEight{}} & \multicolumn{1}{l}{\mistralSeven{}} \\ \midrule
\name{} (Ours) & \textbf{0.79}                      & \textbf{0.84}                  & \textbf{0.78}                  & \textbf{0.79}                    \\
\lats{}        & 0.00                               & 0.00                           & 0.03                           & 0.00                             \\ \bottomrule
\end{tabular}
\end{table}

\subsection{Implementation details of \name{}, ablation studies, and variants}

\begin{table}[htbp]
\scriptsize
    \centering
    \caption{An example of annotated successful trajectory for \alfworld{} in \name{}}
    \ttfamily
    \begin{tabular}{p{0.9\linewidth}}
        \toprule
        You are in the middle of a room. Looking quickly around you, you see a cabinet 4, a cabinet 3, a cabinet 2, a cabinet 1, a countertop 1, a garbagecan 1, a handtowelholder 2, a handtowelholder 1, a sinkbasin 2, a sinkbasin 1, a toilet 1, a toiletpaperhanger 1, and a towelholder 1.

Your task is to: put some spraybottle on toilet.

\textcolor{thought}{> think: To solve the task, I need to find and take a sparybottle, then put it on toilet.}

OK.

\textcolor{critic}{==> critic: The thought is reasonable. This step is GOOD.}

\textcolor{thought}{> think: First I need to find a spraybottle. A spraybottle is more likely to appear in cabinet (1-4), countertop (1), toilet (1), sinkbasin (1-2), garbagecan (1). I can check one by one, starting with cabinet 1.}

OK.

\textcolor{critic}{==> critic: I need to find spraybottle firstly. The thought is reasonable. This step is GOOD.}

> go to cabinet 1

On the cabinet 1, you see a cloth 1, a soapbar 1, a soapbottle 1.

\textcolor{critic}{==> critic: I have not seen spraybottle in this place. This step is UNKNOWN.}

> go to cabinet 2

The cabinet 2 is closed.

\textcolor{critic}{==> critic: The cabinet 2 is closed. This step is UNKNOWN.}

> open cabinet 2

You open the cabinet 2. The cabinet 2 is open. In it, you see a candle 1, and a spraybottle 2.

\textcolor{critic}{==> critic: I have seen spraybottle 2 in this place. This step is GOOD.}

\textcolor{thought}{> think: Now I find a spraybottle (2). Next, I need to take it.}

OK.

\textcolor{critic}{==> critic: I can take the spraybottle after finding it. The thought is reasonable. This step is GOOD.}

> take spraybottle 2 from cabinet 2

You pick up the spraybottle 2 from the cabinet 2.

\textcolor{critic}{==> critic: I have taken spraybottle 2 in this place. This step is GOOD.}

\textcolor{thought}{> think: Now I take a spraybottle (2). Next, I need to put it in/on toilet 1.}

OK.

\textcolor{critic}{==> critic: I can put the spraybottle in/on toilet after taking it. The thought is reasonable. This step is GOOD.}

> go to toilet 1

On the toilet 1, you see a soapbottle 2.

\textcolor{critic}{==> critic: I should go to toilet if I want to put the spraybottle on it. This step is GOOD.}

> put spraybottle 2 in/on toilet 1

You put the spraybottle 2 in/on the toilet 1.

\textcolor{critic}{==> critic: I have put the spraybottle 2 in/on the toilet 1. This step is GOOD.}\\ \bottomrule
    \end{tabular}
    \label{tab:alfworld_ours_example}
\end{table}

\begin{table}[htbp]
\scriptsize
    \centering
    \caption{An example of annotated successful trajectory for \babyai{} in \name{}}
    \ttfamily
    \begin{tabular}{p{0.9\linewidth}}
        \toprule

Goal of the agent: go to the green key

Observation:You see a wall 2 steps right, You see a wall 3 steps forward, You see a critic box 3 steps left, You see a critic key 3 steps left and 1 step forward, You see a blue box 3 steps left and 2 steps forward

\textcolor{thought}{Action:think: To complete the task, I need to find a green key and go to it.}

Observation:OK

\textcolor{critic}{Critic:The thought is reasonable. This step is GOOD.}

\textcolor{thought}{Action:think: I have not seen a green key. Firstly, I need to find a green key.}

Observation:OK

\textcolor{critic}{Critic:I need to find a green key firstly. The thought is reasonable. This step is GOOD.}

Action:turn left

Observation:You see a wall 3 steps right, You see a blue key 3 steps left and 2 steps forward, You see a green key 2 steps left and 1 step forward, You see a green ball 1 step left and 3 steps forward, You see a critic box 3 steps forward, You see a critic key 1 step right and 3 steps forward, You see a blue box 2 steps right and 3 steps forward

\textcolor{critic}{Critic:I have turned left and seen a green key 2 steps left and 1 step forward. This step is GOOD.}

\textcolor{thought}{Action:think: I have seen a green key 2 steps left and 1 step forward. Next, I need to go to the green key.}

Observation:OK

\textcolor{critic}{Critic:I can go to the green key after seeing it. The thought is reasonable. This step is GOOD.}

Action:go forward

Observation:You see a wall 3 steps right, You see a blue key 3 steps left and 1 step forward, You see a green key 2 steps left, You see a green ball 1 step left and 2 steps forward, You see a critic box 2 steps forward, You see a critic key 1 step right and 2 steps forward, You see a blue box 2 steps right and 2 steps forward

\textcolor{critic}{Critic:I have gone forward. The green key is 2 steps left. This step is UNKNOWN.}

Action:turn left

Observation:You see a wall 3 steps left, You see a green box 2 steps left and 3 steps forward, You see a green box 2 steps left and 1 step forward, You see a green key 2 steps forward, You see a blue key 1 step right and 3 steps forward, You see a green ball 2 steps right and 1 step forward, You see a critic box 2 steps right

\textcolor{critic}{Critic:I have turned left. The green key is 2 steps forward. This step is UNKNOWN.}

Action:go forward

Observation:You see a green box 2 steps left and 2 steps forward, You see a green box 2 steps left, You see a green key 1 step forward, You see a blue key 1 step right and 2 steps forward, You see a green ball 2 steps right

\textcolor{critic}{Critic:I have gone forward. The green key is 1 step forward. This step is GOOD.}\\ \bottomrule
    \end{tabular}
    \label{tab:babyai_ours_example}
\end{table}

\begin{table}[htbp]
\scriptsize
    \caption{An example of successful trajectory for \webshop{} in \name{}}
    \centering
    \ttfamily
    \begin{tabular}{p{0.9\linewidth}}
        \toprule
Webshop \\
Instruction:  \\
i would like a 3 ounce bottle of bright citrus deodorant for sensitive skin, and price lower than 50.00 dollars \\
\text{[Search]}  \\
\\
Action: search[3 ounce bright citrus deodorant sensitive skin] \\
Observation: \\
\text{[Back to Search]} \\
Page 1 (Total results: 50) \\
\text{[Next >]} \\
\text{[B078GWRC1J]} \\
Bright Citrus Deodorant by Earth Mama | Natural and Safe for Sensitive Skin, Pregnancy and Breastfeeding, Contains Organic Calendula 3-Ounce \\
\$10.99 \\
\text{[B078GTKVXY]} \\ 
Ginger Fresh Deodorant by Earth Mama | Natural and Safe for Sensitive Skin, Pregnancy and Breastfeeding, Contains Organic Calendula 3-Ounce \\
\$10.99 \\
\text{[B08KBVJ4XN]} \\
Barrel and Oak - Aluminum-Free Deodorant, Deodorant for Men, Essential Oil-Based Scent, 24-Hour Odor Protection, Cedar \& Patchouli Blend, Gentle on Sensitive Skin (Mountain Sage, 2.7 oz, 2-Pack) \\
\$15.95  \\
\textcolor{critic}{Critic: The item required in the instruction is 'bright citrus deodorant for sensitive skin'. I have searched '3 ounce bright citrus deodorant sensitive skin', which is similar to the required item. I also see that [B078GWRC1J] is bright citrus deodorant. This step is GOOD.} \\
\\
Action: click[B078GWRC1J] \\
Observation: \\
\text{[Back to Search]} \\
\text{[< Prev]} \\
scent [assorted scents][bright citrus][calming lavender][ginger fresh][simply non-scents] \\
size [travel set (4-pack)][3 ounce (pack of 1)][3-ounce (2-pack)] \\
Bright Citrus Deodorant by Earth Mama | Natural and Safe for Sensitive Skin, Pregnancy and Breastfeeding, Contains Organic Calendula 3-Ounce \\
Price: \$10.99 \\
Rating: N.A. \\
\text{[Description]} \\
\text{[Features]} \\
\text{[Reviews]} \\
\text{[Buy Now]}  \\
\textcolor{critic}{Critic: I have clicked [B078GWRC1J] and found that this item is 'bright citrus deodorant' and it can satisfy the attributes in instruction, which is '3 ounce' and 'price lower than 50.00 dollars'. This step is GOOD.} \\
\\
Action: click[bright citrus] \\
Observation: You have clicked bright citrus. \\
\textcolor{critic}{Critic: I have clicked [bright cirtrus] to choose the option that satisfies the instruction. This step is GOOD.} \\
\\
Action: click[3 ounce (pack of 1)] \\
Observation: You have clicked 3 ounce (pack of 1). \\
\textcolor{critic}{Critic: I have clicked [3 ounce (pack of 1)] to choose the option that satisfies the instruction. This step is GOOD.} \\
\\
Action: click[Buy Now] \\
Observation: Your score (min 0.0, max 1.0): 1.0 \\ 
\textcolor{critic}{Critic: I have bought the item that satisfies the instruction. This step is GOOD.} \\ \bottomrule
    \end{tabular}
    \label{tab:webshop_ours_example}
\end{table}

We build our method based on \react{} \citep{yao2022react}. Specifically, we adopt the in-context examples from \react{} and annotate reflections for those trajectories. For each task type in \alfworld{}, we annotate two trajectories from \react{} as successful examples to teach LLMs how to complete this type of tasks. For \babyai{}, we annotate two trajectories with thoughts like in \react{}, as well as reflections, for each task type. As to \webshop{}, we annotate the example from \react{} with reflections, and we remove the thoughts annotated in \react{} because we found better results can be achieved without thoughts. The annotated trajectories for \alfworld{}, \babyai{} and \webshop{} are shown in \cref{tab:alfworld_ours_example}, \cref{tab:babyai_ours_example}, and \cref{tab:webshop_ours_example} respectively. The reflections we annotated are labeled in light orange. We also use \react{} algorithm to collect some failed trajectories from the training set for each task type of \alfworld{} and \babyai{}. The collected failed trajectories are also annotated with reflections, aiming to teach LLMs to distinguish between steps that are beneficial, harmful, or indeterminate for achieving a goal.

We annotate the reflections using the following criteria:

\begin{itemize}
    \item[(1)] If an action on a trajectory is necessary for achieving the final goal, the following reflections will end with ``This step is GOOD.'' to indicate the completion of an essential step, such as action ``take spraybottle 2 from cabinet 2'' for the goal ``put some spraybottle on toilet''.
    \item[(2)] If an action is crucial for obtaining essential information to complete a goal, the subsequent reflections will end with ``This step is GOOD.'', such as action ``open cabinet 2'' in \cref{tab:alfworld_ours_example} which finds the target object.
    \item[(3)] If an action is definitely useless or incorrect for achieving a goal, the following reflections will end with ``This step is BAD.'' to indicate this action is undesirable, such as action ``take cloth 1 from cabinet 1'' for the task in \cref{tab:alfworld_ours_example}.
    \item[(4)] If an action can not be evaluated as good or bad from the history, the following reflections will end with ``This step is UNKNOWN.'', such as action ``go to cabinet 1'' or ``go to cabinet 2'' in \cref{tab:alfworld_ours_example}.
\end{itemize}

All the annotated successful and failed trajectories are used to construct fine-tuning examples to fine-tune LLMs to generate better reflections.

Considering the computational cost of fully finetuning of LLMs, we use LoRA \citep{hu2021lora} to finetune our models. In \alfworld{}, with two successful trajectories and one failed trajectory for each task type, we have 485 (input,output) pairs in total of six task types to finetune models. In \babyai{}, the number of finetuning (input,output) pairs is 418. We finetune models for 1,000 steps with learning rate 2.5e-5 and batch size 2. We use A100 GPU with 80GB memory to fine-tune our model. With just about 400-500 (input,output) pairs and 1,000 fine-tuning steps, we can complete the fine-tuning process within one and a half hours.

During testing, the fine-tuned models are used to generate reflections after executing an action in the environment, as well as to forecast the potential outcomes of each action candidate.


The number of candidate actions $n$ is a hyperparameter, set to $5$ in our experiments, that is, we evaluate the top $5$ candidate actions sampled from $\piL$ per state. It is worth noting that sampling top candidate actions introduces a nonzero probability of missing the true argmax action, especially when distinctions among candidate actions are subtle. However, this choice is primarily driven by computational practicality: explicitly computing or evaluating the full action distribution for large or open-ended action spaces common in LLM-based decision-making is typically intractable. Empirically, we find that generating a small subset of candidate actions from a strong LLM prior is often sufficient to include promising actions, thus making the trade-off between computational efficiency and accuracy acceptable.


After sampling action candidates, we use the fine-tuned model to predict future outcomes for each action candidate. The model needs to predict the possible observation and generate reflections for each predicted step. We set the maximum prediction step as 4, the model will continue the prediction until it generates a reflection ending with ``This step is GOOD.'' or ``This step is BAD'', or when it reaches the maximum prediction step.

For the optimization of $\piL$, we solve an optimization problem in \cref{equ:optimization} with a hyper-parameter $\alpha$, which balances the generating probabilities of $\piL$ and the values given by $\gVL$. For \alfworld{}, we set $\alpha$ as 1, which yields superb performance over baselines. For \babyai{}, we conduct a grid-search over \{1/2,1,2,5,10\} for $\alpha$, finding that different LLMs will have best performance with different $\alpha$. The results can be seen in \cref{tab:alpha}. For \webshop{}, we also conduct a grid search over \{1/10,1,10\} to find the best $\alpha$. \cref{equ:optimization} is a weighted combination of the original policy and the action evaluation values. It updates the distribution by adjusting the probabilities of the top candidate actions, while leaving the probabilities of other actions unchanged. 

We set the maximum horizon length to 40 for \alfworld{}, 30 for \babyai{}, and 15 for \webshop{}. If the agent has not reached the final goal after the maximum steps, this episode will be marked as failure.

We use A100 GPU with 80GB memory to evaluate our method. For \name{}, the execution time for \alfworld{} is about 10 hours for 134 tasks using single A100 GPU. And for \babyai{}, the execution time can be varied for different task types, ranging from 4 to 10 hours for 50 tasks using one A100 GPU. For \webshop{}, 100 tasks will be completed within 3 or 4 hours. The GPU memory usage may range from 15GB to over 70GB during the interaction according to the length of inputs to LLMs.

We compare our method with all the aforementioned baselines, demonstrating the effectiveness of our method on decision-making tasks with both high-level actions and low-level actions, even with potentially infinite actions. To demonstrate the effectiveness of each component in our method, we conduct ablation studies on each component. We removes the reflections from \name{} as well as the integration during pre-action-generation phase. This variant is called \namewolangcritic{}. We also evaluate the role of $\gVL$ by removing it from \name{} as well as the integration during post-action-generation phase. This variant is called \namewovaluecritic{}. We also demonstrate the role of the action prior given by LLM policy by using only critic $\gVL$ for decision-making. We call this variant as \nameonlyvaluecritic{}. The execution time of those variants during evaluation can be varied according to its performance because a method having poor performance typically will cost more time to execute. On \alfworld{}, it may be 10-20 hours. The comparisons between those variants are shown in \cref{fig:ablation}.

We found that each component in \name{} is crucial for the superb performance. Removing some components may lead to wrong choice of action candidates. Such an example is shown in \cref{tab:ablation_example}. \name{} can complete this task successfully, while eliminating some components in \name{} will lead to failure. The comparison is shown in \cref{fig:illustration}.

\begin{table}[htbp]
\scriptsize
    \caption{An example of successful trajectory in test environment in \babyai{} given by \name{}}
    \centering
    \ttfamily
    \begin{tabular}{p{0.9\linewidth}}
        \toprule
Goal of the agent:go to the green ball

Observation:You see a wall 5 steps forward, You see a wall 2 steps left, You see a yellow ball 1 step left and 4 steps forward, You see a blue ball 1 step left and 3 steps forward, You see a red ball 1 step left and 1 step forward, You see a grey key 2 steps right and 2 steps forward, You see a green key 2 steps right and 1 step forward, You see a blue key 3 steps right and 1 step forward

\textcolor{thought}{Action:think:To complete the task, I need to find a green ball and go to it.}

Observation:OK

\textcolor{critic}{Critic:This thought is reasonable. This step is GOOD.}

\textcolor{thought}{Action:think:I have seen a green ball 1 step left and 4 steps forward. Next, I need to go to the green ball.}

Observation:OK

\textcolor{critic}{Critic:This thought is reasonable. I can go to the green ball after seeing it. This step is GOOD.}

Action:turn right

Observation:You see a wall 5 steps forward, You see a wall 2 steps right, You see a green ball 3 steps left and 4 steps forward, You see a grey key 2 steps left and 2 steps forward, You see a blue key 1 step left and 3 steps forward, You see a green key 1 step left and 2 steps forward, You see a blue ball 1 step right and 1 step forward

\textcolor{critic}{Critic:I have turned right. The green ball is 3 steps left and 4 steps forward. This step is UNKNOWN.}

Action:go forward

Observation:You see a wall 4 steps forward, You see a green ball 3 steps left and 3 steps forward, You see a grey key 2 steps left and 1 step forward, You see a blue key 1 step left and 2 steps forward, You see a green key 1 step left and 1 step forward, You see a blue ball 1 step right

\textcolor{critic}{Critic:I have gone forward. The green ball is 3 steps left and 3 steps forward. This step is UNKNOWN.}

Action:go forward

Observation:You see a wall 3 steps forward, You see a wall 2 steps right, You see a green ball 3 steps left and 2 steps forward, You see a grey key 2 steps left, You see a blue key 1 step left and 1 step forward, You see a green key 1 step left

\textcolor{critic}{Critic:I have gone forward. The green ball is 3 steps left and 2 steps forward. This step is UNKNOWN.}

Action:go forward

Observation:You see a wall 2 steps forward, You see a wall 2 steps right, You see a green ball 3 steps left and 1 step forward, You see a blue key 1 step left

\textcolor{critic}{Critic:But I have gone forward. The green ball is 3 steps left and 1 step forward. This step is UNKNOWN.}

Action:go forward

Observation:You see a wall 1 step forward, You see a wall 2 steps right, You see a green ball 3 steps left

\textcolor{critic}{Critic:But I have gone forward. The green ball is 3 steps left. This step is UNKNOWN.}

Action:turn left

Observation:You see a wall 1 step right, You see a grey key 2 steps left and 2 steps forward, You see a green key 2 steps left and 1 step forward, You see a blue key 1 step left and 1 step forward, You see a red ball 4 steps forward, You see a green ball 3 steps forward

\textcolor{critic}{Critic:But I have turned left. The green ball is 3 steps forward. This step is UNKNOWN.}

Action:go forward

Observation:You see a wall 1 step right, You see a grey key 2 steps left and 1 step forward, You see a green key 2 steps left, You see a blue key 1 step left, You see a red ball 3 steps forward, You see a green ball 2 steps forward

\textcolor{critic}{Critic:But I have gone forward. The green ball is 2 steps forward. This step is UNKNOWN.}

Action:go forward

Observation:You see a wall 1 step right, You see a grey key 2 steps left, You see a red ball 2 steps forward, You see a green ball 1 step forward

\textcolor{critic}{Critic:But I have gone forward. The green ball is 1 step forward. This step is GOOD.} \\ \bottomrule
    \end{tabular}
    \label{tab:ablation_example}
\end{table}






\end{document}